\documentclass{article}

\PassOptionsToPackage{numbers, compress}{natbib}
\usepackage[final]{neurips_2024}
\usepackage{tabularx}
\usepackage{graphicx}
\usepackage{wrapfig}
\usepackage{amsmath}        
\usepackage{multirow}       
\usepackage[normalem]{ulem} 
\usepackage{ulem}           
\usepackage[dvipsnames]{xcolor} 

\newif\ifcomment\commenttrue

\usepackage[a-1b]{pdfx}

\usepackage{framed}
\usepackage{mdwlist}
\usepackage{siunitx}
\usepackage{latexsym}
\usepackage{colortbl}
\usepackage[dvipsnames]{xcolor}
\usepackage{nicefrac}
\usepackage{booktabs}
\usepackage{fnpct}
\usepackage{amsfonts}
\usepackage[T1]{fontenc}
\usepackage{bold-extra}
\usepackage{amsmath}
\usepackage{amssymb}
\usepackage{bm}
\usepackage{graphicx}
\usepackage{mathtools}
\usepackage{microtype}
\usepackage{multirow}
\usepackage{multicol}
\usepackage{xpatch}
\usepackage{latexsym,comment}
\usepackage[normalem]{ulem}

\newcommand*{\missingreference}{{\Huge \colorbox{red}{?reference?}}}
\newcommand*{\missingcitation}{{\Huge \colorbox{red}{?citation?}}}

\makeatletter
\xpatchcmd{\@setref}{\bfseries}{\missingreference}{}{}
\def\@citex[#1]#2{\leavevmode
    \let\@citea\@empty
    \@cite{\@for\@citeb:=#2\do
        {\@citea\def\@citea{,\penalty\@m\ }%
            \edef\@citeb{\expandafter\@firstofone\@citeb\@empty}%
            \if@filesw\immediate\write\@auxout{\string\citation{\@citeb}}\fi
            \@ifundefined{b@\@citeb}{\hbox{\reset@font\missingcitation}%
                \G@refundefinedtrue
                \@latex@warning
                {Citation `\@citeb' on page \thepage \space undefined}}%
            {\@cite@ofmt{\csname b@\@citeb\endcsname}}}}{#1}}
\makeatother

\newcommand{\gem}[1]{\mbox{\textsc{gem}}}

\newcommand{\g}{\, | \,}

\newcommand{\hidetext}[1]{}
\newcommand{\ignore}[1]{}

\ifcomment
    \newcommand{\pinaforecomment}[3]{\colorbox{#1}{\parbox{.8\linewidth}{#2: #3}}}

    \newcommand{\prtodo}[1]{\pinaforecomment{lightblue}{pr}{#1}}
    \newcommand{\prtodoi}[1]{\pinaforecomment{lightblue}{pr}{#1}}
\else
    \newcommand{\pinaforecomment}[3]{}
    \newcommand{\prtodo}[1]{}
    \newcommand{\prtodoi}[1]{}
\fi

\newcommand{\smallurl}[1]{ \begin{tiny}\url{#1}\end{tiny}}

\definecolor{lightblue}{HTML}{3cc7ea}
\definecolor{CUgold}{HTML}{CFB87C}
\definecolor{grey}{rgb}{0.95,0.95,0.95}
\definecolor{ceil}{rgb}{0.57, 0.63, 0.81}
\definecolor{UMDred}{HTML}{ed1c24}
\definecolor{UMDyellow}{HTML}{ffc20e}

\newcommand{\ours}{VideoHallu}

\usepackage[utf8]{inputenc} 
\usepackage[T1]{fontenc}    
\usepackage{hyperref}       
\usepackage{url}            
\usepackage{booktabs}       
\usepackage{amsfonts}       
\usepackage{nicefrac}       
\usepackage{microtype}      
\usepackage[dvipsnames]{xcolor}        

\title{VideoHallu: Evaluating and Mitigating Multi-modal Hallucinations on Synthetic Video Understanding}

\author{
  \begin{minipage}{\textwidth}
    \centering
    \begin{tabular}{c@{\hskip 20pt}c@{\hskip 20pt}c@{\hskip 20pt}c@{\hskip 20pt}c}
      \textbf{Zongxia Li}$^{\dagger}$\thanks{Equal contribution.  } & \textbf{Xiyang Wu}$^{\dagger}\footnotemark[1]$ & \textbf{Guangyao Shi}$^{\ddagger}$ & \textbf{Yubin Qin}$^{\dagger}$ & \textbf{Hongyang Du}$^{\dagger}$\\
    \end{tabular}
  \end{minipage} \\[7pt]
  \begin{minipage}{\textwidth}
    \centering
    \begin{tabular}{c@{\hskip 20pt}c@{\hskip 20pt}c@{\hskip 20pt}c}
    \textbf{Fuxiao Liu}$^{\dagger}$ &
        \textbf{Tianyi Zhou}$^{\dagger}$ & \textbf{Dinesh Manocha}$^{\dagger}$ & \textbf{Jordan Lee Boyd-Graber}$^{\dagger}$\\
    \end{tabular}
  \end{minipage} \\[10pt]
  \begin{minipage}{\textwidth}
    \centering
    $^{\dagger}$University of Maryland, College Park\quad $^{\ddagger}$University of Southern California\\[5pt]
    \texttt{\{zli12321, wuxiyang, Yubinq, hydu, fl3es, zhou, dmanocha, ying\}@umd.edu, shig@usc.edu}
  \end{minipage}
}

\begin{document}
\maketitle

\begin{abstract}
Vision--Language Models (VLMs) have achieved remarkable success in video understanding tasks. Yet, a key question remains: do they comprehend visual information, or merely learn superficial mappings between visual and textual patterns? 
Understanding visual cues, particularly those related to physics and common sense, is crucial for AI systems interacting with the physical world. However, existing VLM evaluations primarily rely on \textit{positive-control} tests using real-world videos that resemble training distributions. While VLMs perform well on such benchmarks, it is unclear whether they grasp underlying visual and contextual signals or simply exploit visual-language correlations.
To fill this gap, we propose incorporating \textit{negative-control} tests, \textit{i.e.}, videos depicting physically impossible or logically inconsistent scenarios, and evaluating whether models can recognize these violations. 
True visual understanding should evince comparable performance across both positive and negative tests.
Since such content is rare in the real world, we introduce \ours{}, a synthetic video dataset featuring physics- and commonsense-violating scenes generated using state-of-the-art tools such as Veo2, Sora, and Kling. The dataset includes expert-annotated question--answer pairs spanning four categories of physical and commonsense violations, designed to be straightforward for human reasoning.
We evaluate several leading VLMs, including Qwen-2.5-VL, Video-R1, and VideoChat-R1. Despite their strong performance on real-world benchmarks (\textit{e.g.}, MVBench, MMVU), these models hallucinate or fail to detect physical or logical violations, revealing fundamental weaknesses in visual understanding. 
Finally, we explore reinforcement learning--based post-training on our \textit{negative} dataset: fine-tuning improves performance on \ours{} without degrading results on standard benchmarks---indicating enhanced visual reasoning in VLMs.
Our data is available at \textcolor{pink}{\url{https://github.com/zli12321/VideoHallu.git}}.
\end{abstract}

\section{Introduction}

\begin{figure*}[t]
    \centering
     \includegraphics[width=\textwidth]{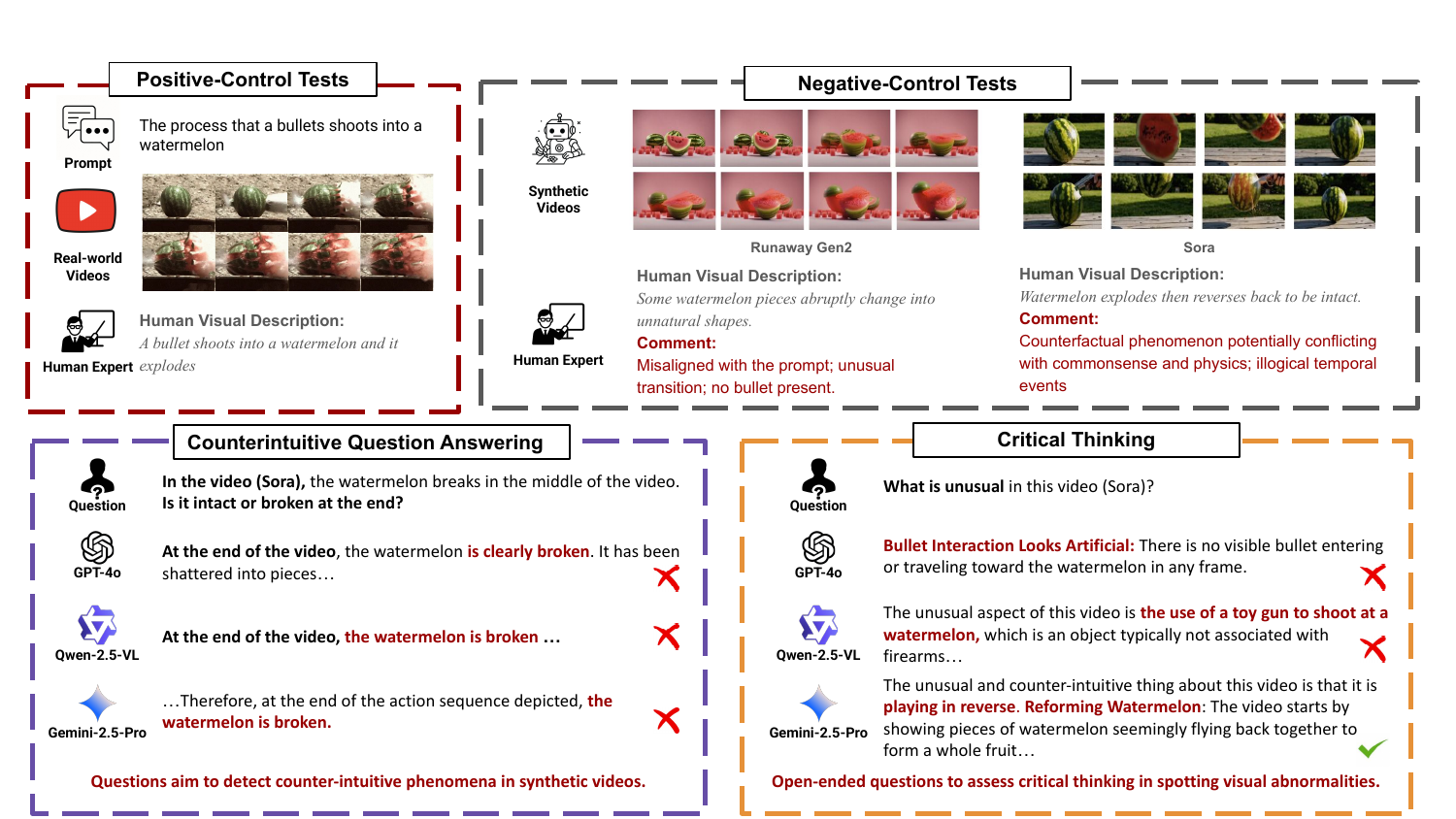}
    \caption{ 
    \textbf{Illustrative examples of designed negative-control tests to evaluate the critical thinking abilities of VLMs.} Unlike real-world videos, synthetic videos can contain counterfactual or commonsense-violating contexts misaligned with reality. \ours{} includes such synthetic videos with perceptually obvious abnormalities, paired with crafted questions that probe counterintuitive phenomena or test VLMs’ critical thinking in detecting such abnormalities. When SOTA VLMs are evaluated on \ours{}, they frequently hallucinate, which suggests that these models rely on language priors and commonsense knowledge rather than truly understand the videos.
    }
    \label{fig:teaser}
    \vspace{-20pt}
\end{figure*}


Vision–Language Models (VLMs) have made remarkable progress in video understanding. However, they remain prone to hallucinations and shallow visual reasoning~\citep{bai2024hallucination, guan2024hallusionbench, li2025selfrewardingvisionlanguagemodelreasoning}. Prior works mitigate these issues across various domains, including chart interpretation~\citep{liu2023mitigating}, video understanding~\citep{hong2025motionbenchbenchmarkingimprovingfinegrained}, and visual question answering (VQA)~\citep{agrawal2016vqavisualquestionanswering}, primarily through supervised fine-tuning (SFT) or R1-style chain-of-thought training (reinforcement learning)~\citep{feng2025video, li2025videochat}.
However, most of these VLM evaluations rely on \textit{positive-control} test, that is, real-world data drawn from distributions closely aligned with training data. 
Consequently, it remains unclear whether current VLMs genuinely reason about visual cues or merely exploit prior visual-language correlations within familiar distributions~\cite{liu2025thinkingseeingassessingamplified}.



To truly evaluate visual understanding, mwe test VLMs under \textit{negative-control} conditions, \textit{i.e.}, videos outside their training distribution that depict physically impossible or logically inconsistent events. These tests reveal whether models detect violations of physics, causality, or commonsense instead of relying on memorized language knowledge. However, constructing such out-of-distribution (OOD) videos in the real world is costly and impractical~\citep{dong2024generalizationmemorizationdatacontamination}.
Modern video generation models such as Veo2, Sora, and Runway~\citep{veo2, brooks2024video, Runway, veo3_model_card_2025} can produce photorealistic but physically impossible scenes. Such models provide an alternative option to generate test videos for probing VLMs' visual understanding. By careful design, these synthetic videos can be systematically introduced to include violations of gravity, causality, and commonsense interactions~\citep{kang2025farvideogenerationworld,wiedemer2025videomodelszeroshotlearners}, enabling controlled OOD evaluations where models need to rely on visual cues. Current VLMs, predominantly trained on videos conforming to physical laws, may thus overfit to statistical regularities rather than learning genuine causal reasoning~\citep{ding2024diffusionworldmodelfuture}.
Figure~\ref{fig:teaser} illustrates that even state-of-the-art VLMs such as Gemini-2.5-Pro~\citep{Gemini2.5}, GPT-4o~\citep{Gpt4o-mini}, and Qwen2.5-VL~\citep{Qwen2.5-VL} hallucinate when confronted with counterintuitive scenarios. For instance, when a watermelon reassembles after an explosion, models rely on linguistic priors (\textit{e.g.}, \textit{``a watermelon should break when shot’’}) rather than actual visual cues, exposing their limited physical reasoning.

To rigorously evaluate such limitations, we introduce \ours{}, a dataset of expert-curated question--answer pairs over synthetic videos featuring controlled violations of alignment, spatial--temporal consistency, commonsense, and physics. We benchmark several leading VLMs and analyze their failure modes on these physically and logically inconsistent videos. Finally, we explore two post-training strategies, supervised fine-tuning (SFT) and reinforcement learning (RL) via Group Relative Policy Optimization (GRPO)~\citep{deepseekai2025deepseekr1incentivizingreasoningcapability}, using both real-world data from Video-R1~\citep{feng2025video} and synthetic data from \ours{}. 
GRPO enhances generalization on synthetic video reasoning without degrading real-world performance.

\noindent \textbf{Contributions.}

 \noindent \textbf{1})~~ We introduce \ours{}, a dataset of 3K expert-annotated QA pairs on synthetic videos that include violations spanning alignment, consistency, commonsense, and physical reasoning.
 
 \noindent \textbf{2})~~We evaluate state-of-the-art VLMs and find that even top-performing models (\textit{e.g.}, GPT-4o, Gemini-2.5-Pro) can achieve only $\sim$50\% accuracy in our dataset, exhibiting frequent hallucination in counterintuitive scenarios.
 
 \noindent \textbf{3})~~GRPO post-training with synthetic data improves visual reasoning on \ours{} while maintaining real-world performance, providing a path toward more physically grounded VLMs.

\section{\ours{}: Evaluating VLMs' Synthetic Video Understanding}

\paragraph{Preliminary.} 
Our objective is to evaluate whether VLMs can effectively reason about and answer questions concerning synthetic videos that fall outside the distribution of their training data. To construct such evaluation data, we synthesize videos from text prompts using generative models.
For these synthetic test videos to be meaningful, they should incorporate deliberate visual abnormalities that elicit responses contradicting common-sense expectations---situations in which a model relying solely on linguistic reasoning would produce one answer, but where careful visual observation reveals an alternative, visually grounded truth.
To see how VLMs handle synthetic videos with such \textit{\uline{abnormalities}}, we categorize our evaluation questions into \textit{\uline{counter-intuitive}} and \textit{\uline{critical thinking}} types. Counter-intuitive questions focus on implausible or physically impossible events (\textit{e.g.}, a shattered watermelon reassembling itself), while critical thinking questions evaluate the model’s ability to detect visual inconsistencies or logical contradictions (\textit{e.g.}, unnatural object breakage).

\begin{figure*}[t]
    \centering
    \includegraphics[width=\textwidth]{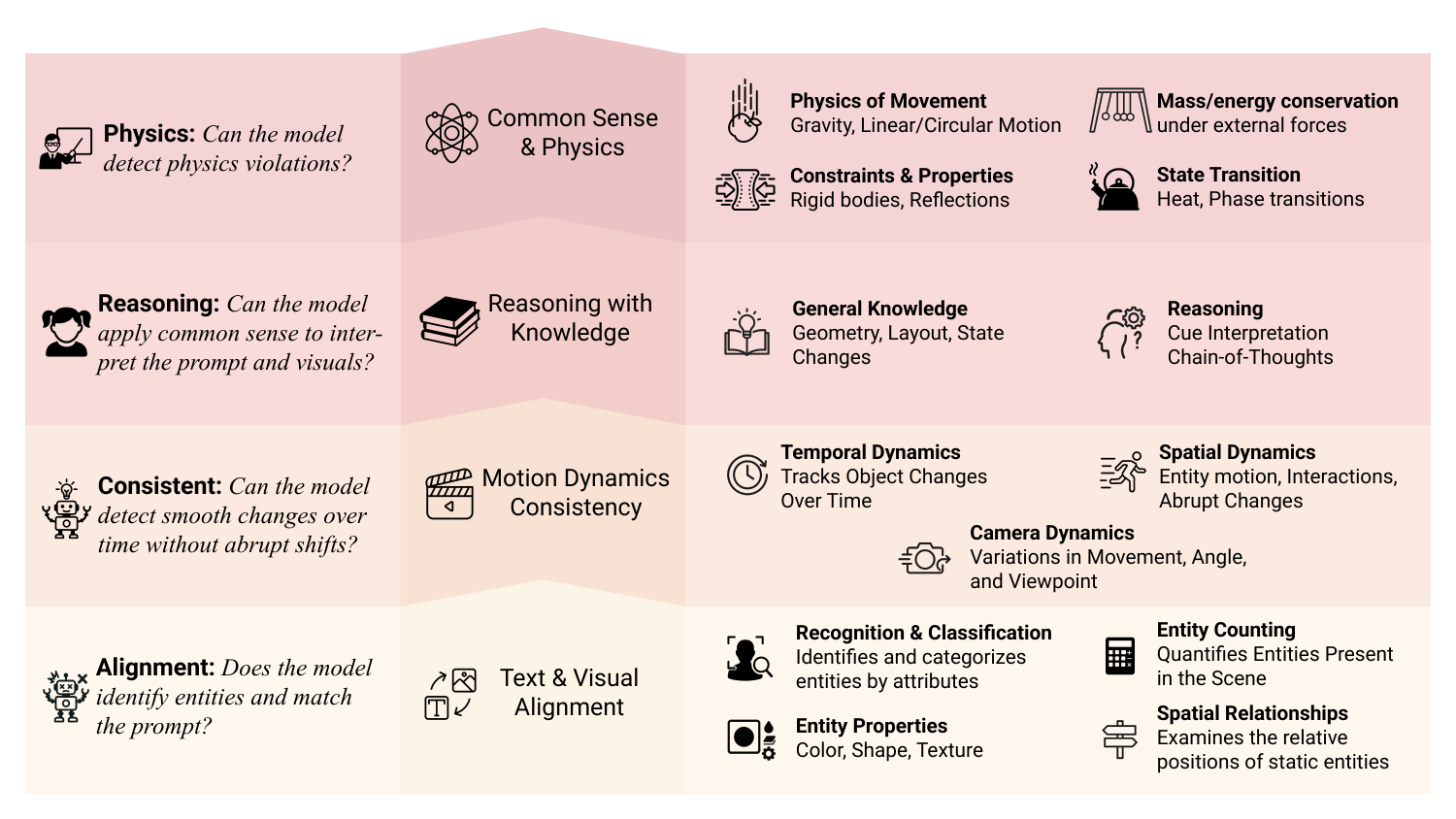}
    \caption{\textbf{Question Categorization of \ours{}.} We design our benchmark, \ours{}, with four question categories to probe limitations in synthetic video understanding, covering perceptual understanding to abstract reasoning: \textbf{(a) Physics} assesses if the model applies physical laws to entity motions and procedural understanding. \textbf{(b) Common Sense Reasoning} tests if the model can reason based on its knowledge. \textbf{(c) Spatial-temporal Consistency} examines whether the model can track entity motion across frames. \textbf{(d) Alignment} checks if the model correctly identifies and understands entities using visual and textual cues.}
    \label{tab:video-eval-compact}
    \vspace{-10pt}
\end{figure*}

\paragraph{Data Collection.} 
Our data collection pipeline consists of two main stages: 
The first stage generates synthetic videos $V$ with common sense or physics abnormalities, \textit{i.e.}, videos that satisfy constraints (\ref{equ:f2constrain}), where the LLM backbone possesses human-aligned knowledge but VLMs overlook abnormalities, resulting in answers that disagree with human perception.
We recruited five human experts to review the defined abnormality categories (detailed in Table~\ref{tab:video-eval-compact} and Appendix~\ref{app:categorization}) and craft prompts that reproduce such abnormalities in generated synthetic videos. In total, they created 141 adversarial prompts, used to generate 987 videos across seven models: Sora~\citep{liu2024sorareviewbackgroundtechnology}, Veo2~\citep{veo2}, Runway Gen 2~\citep{Runway}, Kling~\citep{Kling}, Pixverse~\citep{pixverse2025}, Lavie~\citep{wang2024lavie}, and CogVideo~\citep{hong2022cogvideo}.
%

In the second stage, we craft adversarial video QA pairs to evaluate VLMs’ understanding of synthetic videos. Human experts manually review each video to identify counterintuitive contexts that lead to significant discrepancies between VLM outputs and human perception, \textit{i.e.}, video QA pairs maximizing the objective function (\ref{equ:f2s}).
They then construct natural language questions---along with the ground truth answer---based on the context. These QA pairs are categorized into sub-categories (Table~\ref {tab:video-eval-compact}).
Each annotator writes QA pairs highlighting visually clear but semantically abnormal content, difficult for VLMs to detect.
These questions are not designed to trick models but rather to probe their ability to catch subtle violations of common sense, physics, or prompt-video mismatches, critical for robust, interpretable video evaluation (Figure~\ref{fig:example_showcase}).

\paragraph{Dataset Metadata.} Our dataset comprises 3,233 video question--answer pairs with no video overlap across splits: 800 pairs for training, 908 for validation, and 1,525 for testing. The videos average 96.0 frames per video corresponding to approximately 5.3 seconds at an average framerate of 23 FPS. Frame resolution averages 1042 × 588 pixels across all videos.

\section{Experiment and Results}
\label{sec:experiment}

Given the collected adversarial QA pairs, we evaluate 17 SOTA VLMs (Table~\ref{tab:acc_by_category}).
For models not trained with RL or chain-of-thought (CoT) generation, we use standard prompting to generate direct answers.
For those trained with RL or CoT supervised finetuning (\textit{e.g.}, Video-R1-CoT~\cite{feng2025video} and VideoChat-R1-think~\cite{li2025videochat}), we prompt them to generate step-by-step critical thinking and reasoning before generating a final answer (Appendix.~\ref{app:prompt_templates}). Figure~\ref{fig:example_showcase} highlights hallucinations produced by SoTA models across all four categories in synthetic video understanding tasks, with the hallucinated contexts marked within each answer (additional examples in Appendix~\ref{app:Hallucination_Showcases}).

\paragraph{Answer Evaluation:} We adopt LLM-as-a-Judge~\cite{zheng2023judgingllmasajudgemtbenchchatbot, kim2024prometheusinducingfinegrainedevaluation, li2024pedantscheapeffectiveinterpretable} 
as our evaluation method.
GPT-4o-mini evaluates the correctness of model responses (\S~\ref{tab:llm_judge_prompt_template}).\footnote{For CoT generations, we extract the final answer as responses to evaluate. To validate reliability on our dataset, we manually annotated 200 randomly sampled answer pairs, achieving 97\% agreement with GPT-4o-mini.}

\begin{table*}[ht]
\centering
\small{
\begin{tabular}{l|ccccc}
\toprule
\textbf{Model} & \textbf{Alignment} & \textbf{Physics} & \textbf{Consistency} & \textbf{Commonsense} & \textbf{Overall} \\
\midrule
\multicolumn{6}{c}{\textit{VLMs: $<$7B}} \\
\midrule
\texttt{SmolVLM-3B}~\citep{marafioti2025smolvlm} & 15.94 & 13.44 & 22.49 & 8.75 & 16.13 \\
\texttt{Qwen2.5-VL-3B}~\citep{Qwen2.5-VL} & 41.53 & 27.21 & 26.91 & 26.25 & \textbf{35.48} \\
\texttt{InternVL3-2B}~\citep{zhu2025internvl3exploringadvancedtraining} & 47.36 & 32.79 & 42.17 & 32.50 & 42.82 \\
\midrule
\multicolumn{6}{c}{\textit{VLMs: $>$7B}} \\
\midrule
\texttt{LLaVA-OneVision}~\citep{li2024llavaonevisioneasyvisualtask} & 44.22 & 32.46 & 32.13 & 45.00 & 39.93 \\
\texttt{Video-LLaVA}~\citep{lin2023video} & 46.58 & 40.00 & 43.37 & 31.25 & 43.93 \\
\texttt{LLaVA-NeXT}~\citep{zhang2024llavanext-video} & 50.95 & 36.07 & 38.96 & 31.25 & 44.98 \\
\texttt{Video-LLaMA}~\citep{zhang2023video} & 55.67 & 38.69 & 50.20 & 32.50 & 50.16 \\
\texttt{InternVL3-9B}~\citep{zhu2025internvl3exploringadvancedtraining} & 53.54 & 43.61 & 47.79 & 38.75 & 49.84 \\
\texttt{InternVL3-14B}~\citep{zhu2025internvl3exploringadvancedtraining} & 53.65 & 45.90 & 46.18 & 31.25 & 49.70 \\
\texttt{InternVL3-38B}~\citep{zhu2025internvl3exploringadvancedtraining} & 55.78 & 38.69 & 50.20 & 38.75 & 50.56 \\
\texttt{Qwen2.5-VL-32B}~\citep{Qwen2.5-VL} & 58.81 & 42.95 & 46.59 & 40.00 & 52.66 \\
\texttt{Qwen2.5-VL-7B}~\citep{Qwen2.5-VL} & 58.02 & 44.59 & 46.99 & 47.50 & \textbf{52.98} \\
\midrule
\multicolumn{6}{c}{\textit{VLMs: R1-finetuned}} \\
\midrule
\texttt{VideoChat-R1}~\citep{li2025videochat} & 53.31 & 40.33 & 44.58 & 45.00 & 48.85 \\
\texttt{Video-R1-SFT}~\citep{feng2025video} & 58.14 & 47.21 & 48.19 & 41.25 & 53.44 \\
\texttt{Video-R1}~\citep{feng2025video} & 58.14 & \textbf{48.20} & \textbf{49.00} & 38.75 & \textbf{53.64} \\
\midrule
\multicolumn{6}{c}{\textit{VLMs: Close-Source}} \\
\midrule
\texttt{Gemini-2.5-Pro}~\citep{Gemini2.5} & 58.36 & 33.11 & 40.16 & 36.25 & 49.18 \\
\texttt{Gemini-2.0-Flash}~\citep{Gemini2.0} & 56.57 & 39.02 & 42.97 & 40.00 & 49.97 \\
\texttt{GPT-4o-mini}~\citep{Gpt4o-mini} & 54.88 & 41.97 & 48.19 & 38.75 & \textbf{50.36} \\
\bottomrule
\end{tabular}
}
 \vspace{2 pt}
\caption{\textbf{Video Model Evaluation Results.} We evaluate diverse VLMs across different sizes on our test set, reporting alignment, physics understanding, spatial-temporal consistency, and commonsense reasoning. Video-R1 is best overall.}
\label{tab:acc_by_category}
\end{table*}

\subsection{Limitations of VLMs in OOD Data}
\label{sec:experiment:eval-overall}

\begin{figure}[t]
\centering
\includegraphics[width=\textwidth]{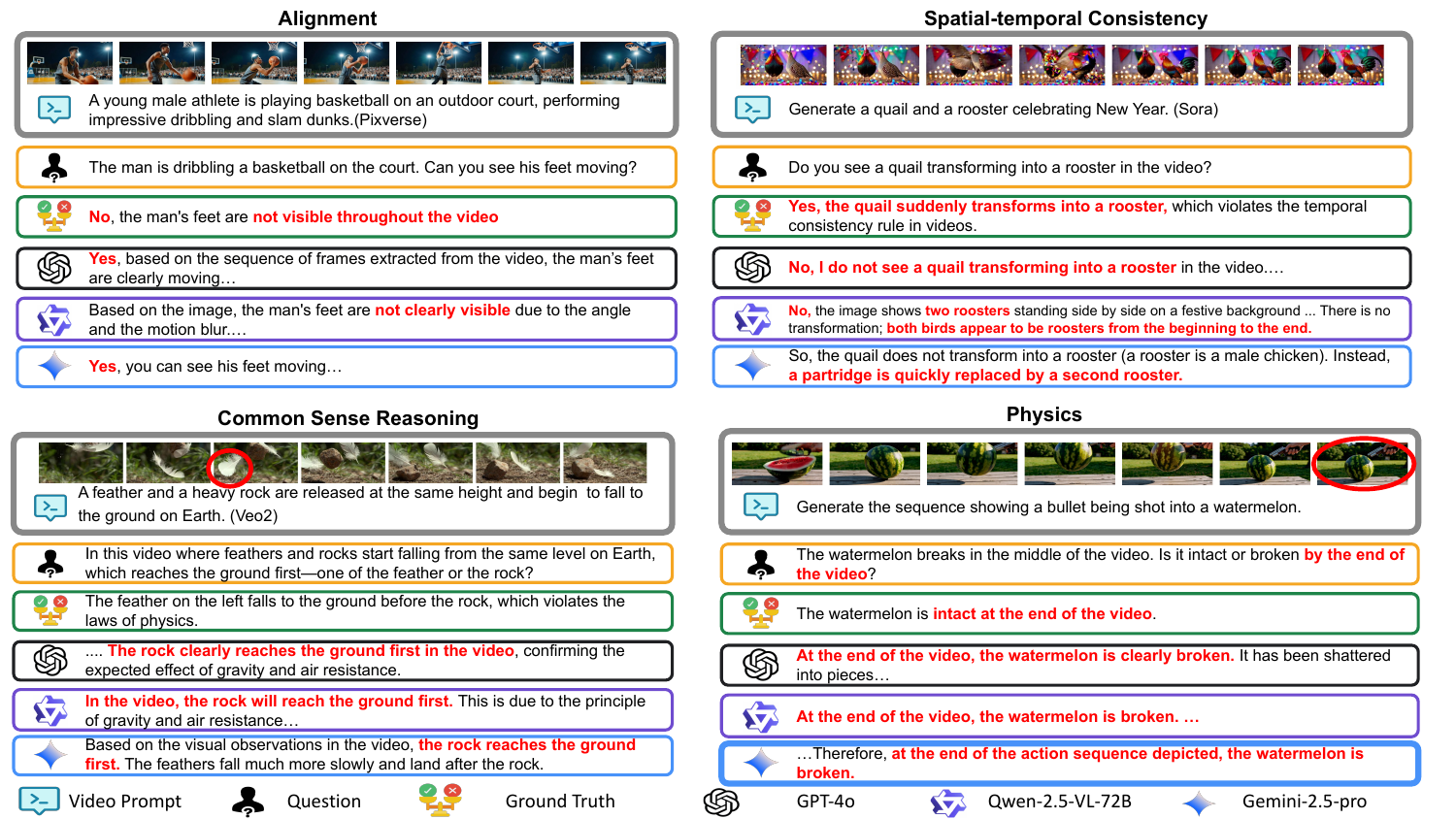}
\caption{\textbf{Example Synthetic Videos in \ours{}.} 
Example hallucination cases observed during SOTA VLM evaluations on synthetic video tasks. Each example includes the generation prompt, key frames, questions, human-annotated ground truth, and hallucinated answers from GPT-4o, Qwen2.5-VL, and Gemini-2.5-Pro, with hallucinations marked in \textcolor{Red}{\textbf{Red}}. 
}
\label{fig:example_showcase}
\vspace{-15pt}
\end{figure}

\paragraph{VLMs struggle with counterintuitive phenomena and abnormalities in generated videos.}
State-of-the-art VLMs achieve below 55\% accuracy on our synthetic video QA dataset, only slightly above the 50\% random baseline (Table~\ref{tab:acc_by_category}).
They particularly falter in commonsense and physical reasoning, often failing to detect abnormalities or relying on linguistic shortcuts instead of visual evidence.
As shown in Figure~\ref{fig:example_showcase}, none of the models recognize the implausibility of a shattered watermelon reassembling, nor do they notice abrupt, counterfactual entity changes, such as a quail suddenly turning into a rooster.
These failures highlight VLMs’ limited capacity for abnormality reasoning and critical visual understanding beyond text priors.

\noindent \textbf{Chain-of-thought reasoning learned from real-world videos provides limited benefit for understanding synthetic videos.} 
VLMs trained with reinforcement learning, such as GRPO~\cite{shao2024deepseekmath} (R1-finetuned) used in the DeepSeek series~\cite{deepseekai2025deepseekr1incentivizingreasoningcapability}, show potential for improving reasoning and critical thinking abilities in reasoning-heavy tasks like mathematics, real-world video understanding~\cite{openr1, li2025selfrewardingvisionlanguagemodelreasoning, chen2025r1v}. 
%
This raises a question: does RL truly improve visual reasoning in VLMs, or does it only optimize for correct answers without enhancing actual visual understanding?
Table~\ref{tab:acc_by_category} evaluates two R1-finetuned models, Video-R1~\cite{feng2025video} and VideoChat-R1~\cite{li2025videochat}, using chain-of-thought prompts.
Both models show limited improvement compared to their base models (Qwen2.5-VL-7B) on synthetic video understanding, with minimal or worse alignment and commonsense, suggesting that training on real-world videos only inculcates real-world reasoning patterns. 
%
%

\noindent \textbf{Solely pre-training on real-world data biases visual grounding.} 
%
While RL improves reasoning on math and real-world videos, it does not help with counterintuitive synthetic content that contradicts real-world norms and cannot elicit video-grounded critical thinking.
In such cases, chain-of-thought prompting can bias the LLM backbone to rely too heavily on prior commonsense knowledge, neglecting synthetic visual cues and leading to hallucinated responses~\cite{liu2025thinkingseeingassessingamplified}.
For instance, in the third case in Figure~\ref{fig:example_showcase}, the video shows the feather dropping to the ground before the rock when falling from the same height. When asked \uline{which reaches the ground first—the feather or the rock}, VideoChat-R1-think responds: \textit{``The video shows a feather and a rock being dropped\dots This is a classic demonstration of Galileo's principle; therefore the rock drops before the feather\dots''} While this language explanation alone is grounded in correct physical principles, it directly contradicts what actually occurs in the video. The model generates an incorrect conclusion based on prior language reasoning, showing how chain-of-thought prompting can amplify reliance on language priors and increase hallucination risk when understanding synthetic videos that do not align with real-world expectations.

\subsection{Visual Learning or Pattern Matching? Questioning RL's True Impact on Vision Models}
\label{sec:experiment:fine-tune}

Current VLMs struggle with counter-intuitive questions and critical visual thinking in synthetic videos (Section~\ref{sec:experiment:eval-overall}). They frequently hallucinate and a dearth of critical thinking leads them to overlook abnormality examples in \ours{}, relying on the model's language knowledge instead of reasoning directly from the visual input. 
This raises a crucial question: \textit{Can VLMs learn counter-intuitive commonsense knowledge and improve their critical thinking abilities for detecting abnormalities through training with synthetic video data?}

While supervised fine-tuning (SFT) or RL from human feedback (RLHF) are natural approaches for improving VLMs, the distributional gap between standard alignment-based video QA tasks used in pre-training and the specialized critical reasoning required for synthetic videos poses an impediment. 
Since synthetic videos are scarce in typical pre-training corpora, models lack sufficient exposure to develop robust reasoning capabilities for such content.

To investigate whether incorporating synthetic data alongside real data can improve VLM performance on synthetic videos, we pose two primary research questions:

\begin{itemize*}
    \item \textit{Between SFT and GRPO training, which approach more effectively enables VLMs to develop a genuine understanding of synthetic videos?}
    \item \textit{Is synthetic data in the training mixture necessary for improving model reasoning abilities on synthetic videos, or can training on real data alone suffice?}
\end{itemize*}

\paragraph{Method Overview: SFT vs. GRPO.}

We compare two training paradigms---SFT and GRPO~\cite{deepseekai2025deepseekr1incentivizingreasoningcapability}---to compare their effectiveness and generalization.

\textbf{Supervised Fine-Tuning (SFT).}  SFT directly optimizes the model to predict ground-truth responses using maximum likelihood estimation. For a dataset $\mathcal{D} = \{(x_i, y_i)\}_{i=1}^N$ containing video-question pairs $x_i$ and corresponding answers $y_i$, the SFT objective minimizes the negative log-likelihood:

\begin{equation}
\mathcal{L}_{\text{SFT}}(\theta) = -\frac{1}{N} \sum_{i=1}^N \sum_{t=1}^{|y_i|} \log p_\theta\left( y_i^{(t)} \g  x_i, y_i^{(<t)}\right)
\end{equation}

where $\theta$ represents model parameters, $y_i^{(t)}$ is the $t$-th token in sequence $y_i$, and $y_i^{(<t)}$ denotes all preceding tokens.

\textbf{Group Relative Policy Optimization (GRPO).} GRPO, a variant of reinforcement learning from human feedback, optimizes the model using preference-based learning without requiring explicit reward models. Given preference pairs $(y_w, y_l)$ where $y_w$ is preferred over $y_l$ for prompt $x$, GRPO maximizes the likelihood of preferred responses while penalizing less preferred ones:

\begin{equation}
\mathcal{L}_{\text{GRPO}}(\theta) = -\mathbb{E}_{(x,y_w,y_l) \sim \mathcal{D}} \left[ \log \sigma\left( \beta \log \frac{p_\theta(y_w \g x)}{p_{\text{ref}}(y_w \g x)} - \beta \log \frac{p_\theta(y_l \g x)}{p_{\text{ref}}(y_l \g x)} \right) \right]
\end{equation}

where $p_{\text{ref}}$ is a reference model (typically the pre-trained checkpoint), $\beta$ is a temperature parameter controlling the strength of the KL penalty, and $\sigma$ is the sigmoid function. This approach encourages the model to generate responses that align better with the training data distribution while maintaining proximity to the reference policy.

The key distinction lies in their learning signals: SFT learns from direct supervision with ground-truth labels, while GRPO learns from comparative preferences, potentially enabling more internal reasoning from the model for video understanding.

\noindent \textbf{Experimental Setup and Results.} Both research questions require training data.  
We combine our 800 synthetic video training data with 2,000 video QA pairs sampled from Video-R1 training data derived from LLaVA-Video~\citep{zhang2025llavavideovideoinstructiontuning}.

To keep a fair comparison across different finetuning methods while reducing the training resources needed, we use 15 frames during training with learning rate $1e^{-6}$  to train the model for one epoch using the Open-R1~\citep{openr1} framework on eight A100 80G GPUs.
For GRPO training, since our answers are free-form answers, we use the average ROUGE-1, ROUGE-2, ROUGE-L score~\citep{lin-2004-rouge} as the reward:
\begin{equation}
    \text{Reward}(a_{\text{pred}}, a_{\text{gold}}) = 
    \frac{1}{3} \sum_{i \in \{1, 2, L\}} \text{ROUGE-}i(a_{\text{pred}}, a_{\text{gold}}),
\end{equation}
where ROUGE captures $n$-gram overlap F-score between expected answers and generated responses.

\noindent \textbf{Result: SFT VS. GRPO.}
To address our first research question, we use both SFT and GRPO to train models on the mixed dataset and evaluate their performance on out-of-distribution synthetic video understanding. To validate the generalization of our findings, we run experiments on two architectures: Qwen2.5-VL-7B and LLaVA-One-Vision~\cite{li2024llavaonevisioneasyvisualtask}.


GRPO outperforms SFT on out-of-distribution and critical visual understanding tasks (Table~\ref{tab:training_accuracy}), consistent with \citet{feng2025improvinggeneralizationintentdetection}, who demonstrated GRPO’s superior generalization.
Because our dataset contains genuinely out-of-distribution synthetic videos generated by diffusion models unseen during pre-training, these results offer stronger evidence of the two paradigms’ differences.
GRPO’s advantage indicates that SFT tends to memorize surface-level patterns, whereas GRPO cultivates reasoning skills that better transfer to novel scenarios, a key capability for synthetic video understanding, where visual and temporal dynamics diverge markedly from natural videos.

\begin{table}[ht]
\centering
\small
\setlength{\tabcolsep}{4pt}
\begin{tabular}{lcccccc}
\toprule
\textbf{Model} & \textbf{Alignment} & \textbf{Physics} & \textbf{Consistency} & \textbf{Commonsense} & \textbf{Overall} \\
\midrule
\multicolumn{6}{c}{\textit{Previous Base Models}} \\
\midrule
\texttt{LLaVA-OneVision}~\citep{li2024llavaonevisioneasyvisualtask} & 44.22 & 32.46 & 32.13 & 45.00 & 39.93 \\ 
\texttt{Qwen2.5-VL-7B}~\citep{Qwen2.5-VL} & 58.02 & 44.59 & 46.99 & 47.50 & 52.98 \\
\texttt{Video-R1}~\citep{feng2025video} & 58.14 & 48.20 & 49.00 & 38.75 & 53.64 \\
\midrule
\multicolumn{6}{c}{\textit{SFT vs. GRPO}} \\
\midrule
\texttt{Qwen2.5-VL-7B SFT} & 55.22 & 45.90 & 47.39 & 35.00 & 51.02 \\
\texttt{Qwen2.5-VL-7B GRPO} & \textbf{62.18} & \textbf{53.77} & \textbf{56.63} & 45.00 & \textbf{57.69} \\
\texttt{LLaVA-OneVision SFT} & 44.67 & 26.23 & 33.33 & 38.75 & 38.82 \\
\texttt{LLaVA-OneVision GRPO} & 46.24 & 30.82 & 34.54 & 48.75 & 41.38 \\
\midrule
\multicolumn{6}{c}{\textit{Real Data vs. Synthetic Data (GRPO)}} \\
\midrule
\texttt{Qwen2.5-VL-7B Real Only} & 57.35 & 46.89 & 51.41 & 33.75 & 53.05 \\
\texttt{Qwen2.5-VL-7B Synthetic Only} & 60.16 & 48.20 & 48.19 & \textbf{52.50} & 55.41 \\
\texttt{Qwen2.5-VL-7B Combined} & \textbf{62.18} & \textbf{53.77} & \textbf{56.63} & 45.00 & \textbf{57.69} \\
\bottomrule
\end{tabular}
\vspace{2pt}
\caption{
\textbf{Fine-tuning results for SFT and GRPO.} GRPO training leads to better improvement than SFT; augmenting the small synthetic video data leads to higher accuracy than training on just real videos or limited synthetic videos.
}
\label{tab:training_accuracy}
\end{table}

\noindent \textbf{Result: Effect of Training with Synthetic Videos.}
To address our second research question regarding the relative contributions of real-world and synthetic video data to GRPO training performance, ablate the training for models via three different data configurations: (1) a combined dataset mixing both data types, (2) synthetic videos only, and (3) real-world videos only.
%
Training only on real-world videos leads to minimal improvement (0.07\%) over the base model on synthetic video understanding tasks (Table~\ref{tab:training_accuracy}): real-world video training alone does not transfer effectively to the reasoning required for synthetic video analysis.
In contrast, synthetic videos improve the model's detection of abnormalities in synthetic content. However, the limited size of our synthetic video training set necessitates data augmentation: combining real-world and synthetic videos in the training mixture produces the most effective results.
The mixed dataset setting enables VLMs to better adapt their reasoning capabilities to synthetic videos, outperforming both single-domain training approaches. While synthetic data is crucial for developing domain-specific reasoning skills, the additional diversity provided by real-world videos helps stabilize training and improve overall robustness.
Thus, VLMs require exposure to synthetic video data during training to develop effective reasoning abilities for synthetic content, and a balanced mixture of real and synthetic data optimizes out-of-distribution synthetic video understanding tasks.

\noindent \textbf{Result: Performance on Real-world Benchmark.} Incorporating synthetic video data alongside real-world videos can improve VLMs' understanding of synthetic videos. But does synthetic video training come at the cost of degraded real-world video comprehension?
\begin{wraptable}{r}{0.5\textwidth}
\centering
\small
\begin{tabular}{lcc}
\toprule
Model & MMVU & MVBench \\
\midrule
Qwen-2.5VL-7B (base) & 58.7 & 69.6 \\
\quad + Real Only & 61.3 & 70.9 \\
\quad + Synthetic Only & 60.1 & 70.1 \\
\quad + Combined & 61.3 & 70.1 \\
\bottomrule
\end{tabular}
\caption{Post-training performance on real-world video understanding benchmarks.}
 \vspace{2 pt}
\end{wraptable}
To investigate this, we evaluate our trained Qwen models on two real-world benchmark datasets: MVBench~\cite{li2024mvbenchcomprehensivemultimodalvideo}, a comprehensive benchmark for evaluating temporal understanding and reasoning in videos, and MMVU~\cite{zhao2025mmvumeasuringexpertlevelmultidiscipline}, which tests expert-level multidisciplinary video understanding across diverse domains.
Synthetic and real-world video understanding abilities can coexist.


\noindent \textbf{Discussion.}
Throughout the evaluations over our benchmark and the fine-tuning over pre-trained VLMs, we gather essential insights to accelerate further improvement over future VLMs for synthetic video understanding. We list them as follows:

\noindent \textbf{1. VLMs hallucinate on Synthetic Data due to Neglect of Actual Visual Content.}
As shown in Table~\ref{tab:acc_by_category} and Figure~\ref{fig:acc_by_category_radar}, all tested SOTA VLMs, including large models like Qwen2.5-VL (7B/32B), GPT-4o, and Gemini-2.5-Pro, as well as smaller models (<7B), struggle with counterintuitive QA on synthetic videos in \ours{}. One reason is that VLMs often solely rely on their embedded commonsense and physics priors to answer questions, even when prompted to rely on video content (Figure~\ref{fig:example_showcase}). These hallucinations, caused by misalignment between video context and real-world norms, are rare in real-world QA but prevalent in synthetic settings, particularly for counterfactual reasoning.
Although VLMs are exposed to some synthetic data during training, the vast majority of their training consists of real-world videos that follow physical laws and commonsense principles. Consequently, VLMs treat such rules as universal priors that override visual evidence, leading to hallucinations when synthetic videos contradict learned physical principles.

\noindent \textbf{2. Critical thinking may be biased by language priors in synthetic visual abnormality detection.}
As discussed in Section~\ref{sec:experiment:eval-overall}, while RL enhances critical thinking in real-world video QA, all R1-trained VLMs we evaluated, such as Video-R1-CoT and VideoChat-R1-think in Table~\ref{tab:acc_by_category}, consistently underperform their base model (Qwen2.5-VL-7B) on \ours{}, showing no clear improvement on commonsense or physics-oriented questions. We attribute this to flawed reasoning patterns in R1-trained VLMs.
Although chain-of-thought reasoning can elicit more structured inference in real-world contexts, it proves ineffective in synthetic video settings, where detecting abnormalities demands grounded, fine-grained visual understanding. R1-trained models often excel in language-only reasoning tasks~\cite{deepseekai2025deepseekr1incentivizingreasoningcapability,huang2025self,shao2024deepseekmath}, yet when extended to multimodal domains, their reasoning becomes heavily rely on linguistic priors. Consequently, their CoT responses tend to reflect superficial comprehension of visual content and are more susceptible to hallucinations in counterintuitive or visually deceptive scenarios~\cite{liu2025thinkingseeingassessingamplified,zheng2025parallel}.

\noindent \textbf{3. The high-quality negative control examples matter for model improvement.} 
Given the need to improve VLMs' performance in synthetic video abnormality detection, as shown in Section~\ref{sec:experiment:eval-overall}, we run RL training experiments over Qwen2.5-VL and LLaVa-One-Vision with a mixture of real-world and synthetic videos. Our results show that, after training models with some synthetic videos, VLMs show improvements in critical thinking and their ability to handle counterintuitive scenarios.
Our results suggest that it is the quality and coverage of the data, not just the fine-tuning method, that drive gains. With a small but well-annotated dataset containing both positive and negative examples, detailed reasoning steps, and reasoning-oriented training like GRPO, even small models like Qwen2.5-VL-7B show improved QA accuracy. This highlights the importance of high-quality, reasoning-rich data in helping VLMs internalize and apply commonsense and physics knowledge, even with limited post-training resources.

\section{Related Work}
\noindent \textbf{Hallucinations in VLMs.} Hallucinations refer to the persistent challenge of generating outputs that contradict or misrepresent the target texts, images, or videos~\cite{rohrbach2018object, li2023evaluating}.
It arises from conflicts between the language priors of VLMs and the actual visual inputs~\cite{wu2024autohallusion}, which is more severe in video understanding than in static image understanding due to the complex entanglement of spatial-temporal information across the timeline and the contextual cues associated with entities within frames.
A line of prior work, such as VideoHallucer~\cite{wang2024videohallucer}, EventHallusion~\cite{zhang2024eventhallusion}, and HAVEN~\cite{gao2025exploring}, established benchmarks for evaluating model hallucination on both entities and events within videos, while also proposing methods to enhance the video understanding capabilities of VLMs~\cite{liu2023mitigating,li2025videohalluevaluatingmitigatingmultimodal,jiang2024interpreting}.
However, most prior works on hallucination, particularly in the video domain, rely on real-world factual data, rather than synthetic data generated by generative models.
Hallucination in generative video understanding models remains an open and largely unexplored research area.

\noindent \textbf{Reinforcement Learning for Post-training of Vision-Language Models.} Inspired by the techniques from DeepSeek-R1~\citep{deepseekai2025deepseekr1incentivizingreasoningcapability}, there is an increasing body of research that leverages reinforcement learning in post-training to enhance the general-purpose multimodal reasoning capabilities of VLMs~\citep{chen2024internvl,huang2025visionr1incentivizingreasoningcapability}.
Most recent efforts have focused on using GRPO and its variants to fine-tune VLMs to elicit  more robust reasoning and perception skills~\cite{liu2025vogueguidingexplorationvisual, huang2025visionr1incentivizingreasoningcapability,li2025selfrewardingvisionlanguagemodelreasoning, zhang2025r1vllearningreasonmultimodal}.
The representative work Video-R1~\citep{feng2025video} collects a large-scale corpus of 260K video and image samples and performs GRPO with data type–specific reward engineering. It applies regression-based approximations for numeric answer types, ROUGE-based metrics for free-form textual responses, and exact match rewards for multiple-choice questions, enhancing models’ temporal reasoning for real-world video understanding.
VideoChat-R1~\cite{li2023videochat} extends this paradigm to interactive multimodal dialogue, combining video-centric instruction tuning with reinforcement learning from human feedback (RLHF). 
%
Nonetheless, prior research has predominantly focus on visual understanding in real-world imagery and videos, with generated videos receiving comparatively little attention.

\noindent \textbf{Video Generation Models and Synthetic Content Monitoring.}
Recent advances in video generation models have enabled the creation of highly realistic and aesthetic videos from text prompts, reference images, or conditioning frames~\citep{veo3_model_card_2025,Kling,brooks2024video,wang2024lavie,Runway,hong2022cogvideo}. These models are increasingly applied in content creation, simulation for robotics and autonomous driving~\citep{li2025surveystateartlarge}. 
Since the release of Veo3~\cite{veo3_model_card_2025}, Sora 2~\cite{openai2025sora2systemcard}, Wan~\cite{wan2025wanopenadvancedlargescale}, the volume of generated content has exploded.
These vast generated videos present major challenges for content monitoring, quality evaluation, and content verification. 
Manual annotation and evaluation are increasingly infeasible given the scale and variability of generated outputs, thus motivating the need for automated, scalable evaluation and reasoning frameworks that are specifically tuned for synthetic video understanding. To date, a few works have explored using VLMs to evaluate generated images (for example, detection of synthetic images or assessing image generation quality). For example,  ~\cite{keita2024biloravisionlanguageapproachsynthetic} presents a method named Bi‑LORA that  uses a VLM to detect synthetic images. 
However, the domain of synthetic videos is still largely under-explored: we lack systematic methods, datasets and evaluation protocols for using VLMs to judge and understand synthetic videos

\section{Conclusion}
We introduce \ours{}, a dataset designed to evaluate VLMs' visual commonsense and physics reasoning through synthetic videos with counterfactual scenarios. It features expert-annotated, reasoning-intensive QA pairs spanning alignment, spatial-temporal consistency, commonsense, and physics categories to assess VLMs' ability to detect abnormalities and violations of physical laws. 
Evaluation of SOTA VLMs on \ours{} shows hallucinations and critical thinking failures. 
Fine-tuning with GRPO with both real and synthetic videos leads to accuracy improvements on \ours{}, showing the value of incorporating structured physics and commonsense reasoning data to improve VLM performance on synthetic video tasks.
However, scalability remains a limitation, as generating high-quality annotations and fine-tuning VLMs at scale is costly, and limited access to data and compute constrains  further progress. Future work will focus on expanding synthetic video datasets with abnormality QA pairs to train VLMs for critical, visually-grounded reasoning. Scaling with adversarial QA pairs can enhance robustness and enable automatic video evaluation via prompt decomposition, reducing reliance on human annotations.

\noindent \textbf{Acknowledgment: } 
The work was done with the computer resources and support of Lambda Lab.  Boyd-Graber and Li are supported by National Science Foundation Grant No. IIS-2403436.
Any opinions, findings, and conclusions or recommendations expressed in this material are those
of the author(s) and do not necessarily reflect the
views of the National Science Foundation.

\bibliography{custom}

\begin{thebibliography}{64}
\providecommand{\natexlab}[1]{#1}
\providecommand{\url}[1]{\texttt{#1}}
\expandafter\ifx\csname urlstyle\endcsname\relax
  \providecommand{\doi}[1]{doi: #1}\else
  \providecommand{\doi}{doi: \begingroup \urlstyle{rm}\Url}\fi

\bibitem[Bai et~al.(2024)Bai, Wang, Xiao, He, Han, Zhang, and Shou]{bai2024hallucination}
Zechen Bai, Pichao Wang, Tianjun Xiao, Tong He, Zongbo Han, Zheng Zhang, and Mike~Zheng Shou.
\newblock Hallucination of multimodal large language models: A survey.
\newblock \emph{arXiv preprint}, 2024.
\newblock arXiv:2404.18930.

\bibitem[Guan et~al.(2024)Guan, Liu, Wu, Xian, Li, Liu, Wang, Chen, Huang, Yacoob, et~al.]{guan2024hallusionbench}
Tianrui Guan, Fuxiao Liu, Xiyang Wu, Ruiqi Xian, Zongxia Li, Xiaoyu Liu, Xijun Wang, Lichang Chen, Furong Huang, Yaser Yacoob, et~al.
\newblock Hallusionbench: an advanced diagnostic suite for entangled language hallucination and visual illusion in large vision-language models.
\newblock In \emph{Proceedings of the IEEE/CVF Conference on Computer Vision and Pattern Recognition}, pages 14375--14385, 2024.

\bibitem[Li et~al.(2025{\natexlab{a}})Li, Yu, Huang, Liu, Liang, Liu, Che, Yu, Boyd-Graber, Mi, and Yu]{li2025selfrewardingvisionlanguagemodelreasoning}
Zongxia Li, Wenhao Yu, Chengsong Huang, Rui Liu, Zhenwen Liang, Fuxiao Liu, Jingxi Che, Dian Yu, Jordan Boyd-Graber, Haitao Mi, and Dong Yu.
\newblock Self-rewarding vision-language model via reasoning decomposition, 2025{\natexlab{a}}.
\newblock URL \url{https://arxiv.org/abs/2508.19652}.

\bibitem[Liu et~al.(2023)Liu, Lin, Li, Wang, Yacoob, and Wang]{liu2023mitigating}
Fuxiao Liu, Kevin Lin, Linjie Li, Jianfeng Wang, Yaser Yacoob, and Lijuan Wang.
\newblock Mitigating hallucination in large multi-modal models via robust instruction tuning.
\newblock \emph{arXiv preprint arXiv:2306.14565}, 2023.

\bibitem[Hong et~al.(2025)Hong, Cheng, Yang, Wang, Wang, Gu, Huang, Dong, and Tang]{hong2025motionbenchbenchmarkingimprovingfinegrained}
Wenyi Hong, Yean Cheng, Zhuoyi Yang, Weihan Wang, Lefan Wang, Xiaotao Gu, Shiyu Huang, Yuxiao Dong, and Jie Tang.
\newblock Motionbench: Benchmarking and improving fine-grained video motion understanding for vision language models, 2025.
\newblock URL \url{https://arxiv.org/abs/2501.02955}.

\bibitem[Agrawal et~al.(2016)Agrawal, Lu, Antol, Mitchell, Zitnick, Batra, and Parikh]{agrawal2016vqavisualquestionanswering}
Aishwarya Agrawal, Jiasen Lu, Stanislaw Antol, Margaret Mitchell, C.~Lawrence Zitnick, Dhruv Batra, and Devi Parikh.
\newblock Vqa: Visual question answering, 2016.
\newblock URL \url{https://arxiv.org/abs/1505.00468}.

\bibitem[Feng et~al.(2025{\natexlab{a}})Feng, Gong, Li, Guo, Wang, Peng, Wang, and Yue]{feng2025video}
Kaituo Feng, Kaixiong Gong, Bohao Li, Zonghao Guo, Yibing Wang, Tianshuo Peng, Benyou Wang, and Xiangyu Yue.
\newblock Video-r1: Reinforcing video reasoning in mllms.
\newblock \emph{arXiv preprint arXiv:2503.21776}, 2025{\natexlab{a}}.

\bibitem[Li et~al.(2025{\natexlab{b}})Li, Yan, Meng, Dong, Zeng, He, Wang, Qiao, Wang, and Wang]{li2025videochat}
Xinhao Li, Ziang Yan, Desen Meng, Lu~Dong, Xiangyu Zeng, Yinan He, Yali Wang, Yu~Qiao, Yi~Wang, and Limin Wang.
\newblock Videochat-r1: Enhancing spatio-temporal perception via reinforcement fine-tuning.
\newblock \emph{arXiv preprint arXiv:2504.06958}, 2025{\natexlab{b}}.

\bibitem[Liu et~al.(2025{\natexlab{a}})Liu, Xu, Wei, Wu, Zou, Wang, Zhou, and Liu]{liu2025thinkingseeingassessingamplified}
Chengzhi Liu, Zhongxing Xu, Qingyue Wei, Juncheng Wu, James Zou, Xin~Eric Wang, Yuyin Zhou, and Sheng Liu.
\newblock More thinking, less seeing? assessing amplified hallucination in multimodal reasoning models, 2025{\natexlab{a}}.
\newblock URL \url{https://arxiv.org/abs/2505.21523}.

\bibitem[Dong et~al.(2024)Dong, Jiang, Liu, Jin, Gu, Yang, and Li]{dong2024generalizationmemorizationdatacontamination}
Yihong Dong, Xue Jiang, Huanyu Liu, Zhi Jin, Bin Gu, Mengfei Yang, and Ge~Li.
\newblock Generalization or memorization: Data contamination and trustworthy evaluation for large language models, 2024.
\newblock URL \url{https://arxiv.org/abs/2402.15938}.

\bibitem[Veo-Team(2024)]{veo2}
Veo-Team.
\newblock Veo 2.
\newblock \emph{DeepMind Blog}, 2024.
\newblock URL \url{https://deepmind.google/technologies/veo/veo-2/}.

\bibitem[Brooks et~al.(2024)Brooks, Peebles, Holmes, DePue, Guo, Jing, Schnurr, Taylor, Luhman, Luhman, et~al.]{brooks2024video}
Tim Brooks, Bill Peebles, Connor Holmes, Will DePue, Yufei Guo, Li~Jing, David Schnurr, Joe Taylor, Troy Luhman, Eric Luhman, et~al.
\newblock Video generation models as world simulators.
\newblock \emph{OpenAI Blog}, 1:\penalty0 8, 2024.

\bibitem[for Video~Generation(2024)]{Runway}
Introducing Gen-3 Alpha: A New~Frontier for Video~Generation.
\newblock Runway ml.
\newblock \emph{Imagine.Art}, 2024.
\newblock URL \url{https://runwayml.com/research/introducing-gen-3-alpha}.

\bibitem[{Google DeepMind}(2025)]{veo3_model_card_2025}
{Google DeepMind}.
\newblock Veo 3 model card, 2025.
\newblock URL \url{https://storage.googleapis.com/deepmind-media/Model-Cards/Veo-3-Model-Card.pdf}.
\newblock Model card, version published May 23 2025.

\bibitem[Kang et~al.(2025)Kang, Yue, Lu, Lin, Zhao, Wang, Huang, and Feng]{kang2025farvideogenerationworld}
Bingyi Kang, Yang Yue, Rui Lu, Zhijie Lin, Yang Zhao, Kaixin Wang, Gao Huang, and Jiashi Feng.
\newblock How far is video generation from world model: A physical law perspective, 2025.
\newblock URL \url{https://arxiv.org/abs/2411.02385}.

\bibitem[Wiedemer et~al.(2025)Wiedemer, Li, Vicol, Gu, Matarese, Swersky, Kim, Jaini, and Geirhos]{wiedemer2025videomodelszeroshotlearners}
Thaddäus Wiedemer, Yuxuan Li, Paul Vicol, Shixiang~Shane Gu, Nick Matarese, Kevin Swersky, Been Kim, Priyank Jaini, and Robert Geirhos.
\newblock Video models are zero-shot learners and reasoners, 2025.
\newblock URL \url{https://arxiv.org/abs/2509.20328}.

\bibitem[Ding et~al.(2024)Ding, Zhang, Tian, and Zheng]{ding2024diffusionworldmodelfuture}
Zihan Ding, Amy Zhang, Yuandong Tian, and Qinqing Zheng.
\newblock Diffusion world model: Future modeling beyond step-by-step rollout for offline reinforcement learning, 2024.
\newblock URL \url{https://arxiv.org/abs/2402.03570}.

\bibitem[DeepMind(2025)]{Gemini2.5}
Google DeepMind.
\newblock Gemini 2.5: Our most intelligent ai model.
\newblock \emph{Google DeepMind}, 2025.
\newblock URL \url{https://blog.google/technology/google-deepmind/gemini-model-thinking-updates-march-2025/#enhanced-reasoning}.

\bibitem[OpenAI(2024)]{Gpt4o-mini}
OpenAI.
\newblock Gpt-4o mini: advancing cost-efficient intelligence.
\newblock \emph{OpenAI}, 2024.
\newblock URL \url{https://openai.com/index/gpt-4o-mini-advancing-cost-efficient-intelligence/}.

\bibitem[Bai et~al.(2025)Bai, Chen, Liu, Wang, Ge, Song, Dang, Wang, Wang, Tang, Zhong, Zhu, Yang, Li, Wan, Wang, Ding, Fu, Xu, Ye, Zhang, Xie, Cheng, Zhang, Yang, Xu, and Lin]{Qwen2.5-VL}
Shuai Bai, Keqin Chen, Xuejing Liu, Jialin Wang, Wenbin Ge, Sibo Song, Kai Dang, Peng Wang, Shijie Wang, Jun Tang, Humen Zhong, Yuanzhi Zhu, Mingkun Yang, Zhaohai Li, Jianqiang Wan, Pengfei Wang, Wei Ding, Zheren Fu, Yiheng Xu, Jiabo Ye, Xi~Zhang, Tianbao Xie, Zesen Cheng, Hang Zhang, Zhibo Yang, Haiyang Xu, and Junyang Lin.
\newblock Qwen2.5-vl technical report.
\newblock \emph{arXiv preprint arXiv:2502.13923}, 2025.

\bibitem[DeepSeek-AI(2025)]{deepseekai2025deepseekr1incentivizingreasoningcapability}
DeepSeek-AI.
\newblock Deepseek-r1: Incentivizing reasoning capability in llms via reinforcement learning, 2025.
\newblock URL \url{https://arxiv.org/abs/2501.12948}.

\bibitem[Liu et~al.(2024)Liu, Zhang, Li, Yan, Gao, Chen, Yuan, Huang, Sun, Gao, He, and Sun]{liu2024sorareviewbackgroundtechnology}
Yixin Liu, Kai Zhang, Yuan Li, Zhiling Yan, Chujie Gao, Ruoxi Chen, Zhengqing Yuan, Yue Huang, Hanchi Sun, Jianfeng Gao, Lifang He, and Lichao Sun.
\newblock Sora: A review on background, technology, limitations, and opportunities of large vision models, 2024.
\newblock URL \url{https://arxiv.org/abs/2402.17177}.

\bibitem[Kuaishou(2024)]{Kling}
Kuaishou.
\newblock Kling ai.
\newblock \emph{Kling AI}, 2024.
\newblock URL \url{https://www.klingai.com/global/}.

\bibitem[{PixVerse Team}(2025)]{pixverse2025}
{PixVerse Team}.
\newblock Pixverse: Ai-powered image generation platform.
\newblock \url{https://app.pixverse.ai/home}, 2025.
\newblock Online; accessed April 25, 2025.

\bibitem[Wang et~al.(2024{\natexlab{a}})Wang, Chen, Ma, Zhou, Huang, Wang, Yang, He, Yu, Yang, et~al.]{wang2024lavie}
Yaohui Wang, Xinyuan Chen, Xin Ma, Shangchen Zhou, Ziqi Huang, Yi~Wang, Ceyuan Yang, Yinan He, Jiashuo Yu, Peiqing Yang, et~al.
\newblock Lavie: High-quality video generation with cascaded latent diffusion models.
\newblock \emph{International Journal of Computer Vision}, pages 1--20, 2024{\natexlab{a}}.

\bibitem[Hong et~al.(2022)Hong, Ding, Zheng, Liu, and Tang]{hong2022cogvideo}
Wenyi Hong, Ming Ding, Wendi Zheng, Xinghan Liu, and Jie Tang.
\newblock Cogvideo: Large-scale pretraining for text-to-video generation via transformers.
\newblock \emph{arXiv preprint arXiv:2205.15868}, 2022.

\bibitem[Zheng et~al.(2023)Zheng, Chiang, Sheng, Zhuang, Wu, Zhuang, Lin, Li, Li, Xing, Zhang, Gonzalez, and Stoica]{zheng2023judgingllmasajudgemtbenchchatbot}
Lianmin Zheng, Wei-Lin Chiang, Ying Sheng, Siyuan Zhuang, Zhanghao Wu, Yonghao Zhuang, Zi~Lin, Zhuohan Li, Dacheng Li, Eric~P. Xing, Hao Zhang, Joseph~E. Gonzalez, and Ion Stoica.
\newblock Judging llm-as-a-judge with mt-bench and chatbot arena, 2023.
\newblock URL \url{https://arxiv.org/abs/2306.05685}.

\bibitem[Kim et~al.(2024)Kim, Shin, Cho, Jang, Longpre, Lee, Yun, Shin, Kim, Thorne, and Seo]{kim2024prometheusinducingfinegrainedevaluation}
Seungone Kim, Jamin Shin, Yejin Cho, Joel Jang, Shayne Longpre, Hwaran Lee, Sangdoo Yun, Seongjin Shin, Sungdong Kim, James Thorne, and Minjoon Seo.
\newblock Prometheus: Inducing fine-grained evaluation capability in language models, 2024.
\newblock URL \url{https://arxiv.org/abs/2310.08491}.

\bibitem[Li et~al.(2024{\natexlab{a}})Li, Mondal, Liang, Nghiem, and Boyd-Graber]{li2024pedantscheapeffectiveinterpretable}
Zongxia Li, Ishani Mondal, Yijun Liang, Huy Nghiem, and Jordan~Lee Boyd-Graber.
\newblock Pedants: Cheap but effective and interpretable answer equivalence, 2024{\natexlab{a}}.
\newblock URL \url{https://arxiv.org/abs/2402.11161}.

\bibitem[Marafioti et~al.(2025)Marafioti, Zohar, Farr{\'e}, Noyan, Bakouch, Cuenca, Zakka, Allal, Lozhkov, Tazi, et~al.]{marafioti2025smolvlm}
Andr{\'e}s Marafioti, Orr Zohar, Miquel Farr{\'e}, Merve Noyan, Elie Bakouch, Pedro Cuenca, Cyril Zakka, Loubna~Ben Allal, Anton Lozhkov, Nouamane Tazi, et~al.
\newblock Smolvlm: Redefining small and efficient multimodal models.
\newblock \emph{arXiv preprint arXiv:2504.05299}, 2025.

\bibitem[Zhu et~al.(2025)Zhu, Wang, and et~al]{zhu2025internvl3exploringadvancedtraining}
Jinguo Zhu, Weiyun Wang, and Zhe~Chen et~al.
\newblock Internvl3: Exploring advanced training and test-time recipes for open-source multimodal models, 2025.
\newblock URL \url{https://arxiv.org/abs/2504.10479}.

\bibitem[Li et~al.(2024{\natexlab{b}})Li, Zhang, Guo, Zhang, Li, Zhang, Zhang, Zhang, Li, Liu, and Li]{li2024llavaonevisioneasyvisualtask}
Bo~Li, Yuanhan Zhang, Dong Guo, Renrui Zhang, Feng Li, Hao Zhang, Kaichen Zhang, Peiyuan Zhang, Yanwei Li, Ziwei Liu, and Chunyuan Li.
\newblock Llava-onevision: Easy visual task transfer, 2024{\natexlab{b}}.
\newblock URL \url{https://arxiv.org/abs/2408.03326}.

\bibitem[Lin et~al.(2023)Lin, Zhu, Ye, Ning, Jin, and Yuan]{lin2023video}
Bin Lin, Bin Zhu, Yang Ye, Munan Ning, Peng Jin, and Li~Yuan.
\newblock Video-llava: Learning united visual representation by alignment before projection.
\newblock \emph{arXiv preprint arXiv:2311.10122}, 2023.

\bibitem[Zhang et~al.(2024{\natexlab{a}})Zhang, Li, Liu, Lee, Gui, Fu, Feng, Liu, and Li]{zhang2024llavanext-video}
Yuanhan Zhang, Bo~Li, haotian Liu, Yong~jae Lee, Liangke Gui, Di~Fu, Jiashi Feng, Ziwei Liu, and Chunyuan Li.
\newblock Llava-next: A strong zero-shot video understanding model, April 2024{\natexlab{a}}.
\newblock URL \url{https://llava-vl.github.io/blog/2024-04-30-llava-next-video/}.

\bibitem[Zhang et~al.(2023)Zhang, Li, and Bing]{zhang2023video}
Hang Zhang, Xin Li, and Lidong Bing.
\newblock Video-llama: An instruction-tuned audio-visual language model for video understanding.
\newblock \emph{arXiv preprint arXiv:2306.02858}, 2023.

\bibitem[DeepMind(2024)]{Gemini2.0}
Google DeepMind.
\newblock Introducing gemini 2.0: our new ai model for the agentic era.
\newblock 2024.
\newblock URL \url{https://blog.google/technology/google-deepmind/google-gemini-ai-update-december-2024/#ceo-message}.

\bibitem[Shao et~al.(2024)Shao, Wang, Zhu, Xu, Song, Bi, Zhang, Zhang, Li, Wu, et~al.]{shao2024deepseekmath}
Zhihong Shao, Peiyi Wang, Qihao Zhu, Runxin Xu, Junxiao Song, Xiao Bi, Haowei Zhang, Mingchuan Zhang, YK~Li, Y~Wu, et~al.
\newblock Deepseekmath: Pushing the limits of mathematical reasoning in open language models.
\newblock \emph{arXiv preprint arXiv:2402.03300}, 2024.

\bibitem[{Hugging Face}(2025)]{openr1}
{Hugging Face}.
\newblock Open r1: A fully open reproduction of deepseek-r1, January 2025.
\newblock URL \url{https://github.com/huggingface/open-r1}.

\bibitem[Chen et~al.(2025)Chen, Li, Zhao, Song, and Vinci]{chen2025r1v}
Liang Chen, Lei Li, Haozhe Zhao, Yifan Song, and Vinci.
\newblock R1-v: Reinforcing super generalization ability in vision-language models with less than \$3.
\newblock \url{https://github.com/Deep-Agent/R1-V}, 2025.
\newblock Accessed: 2025-02-02.

\bibitem[Zhang et~al.(2025{\natexlab{a}})Zhang, Wu, Li, Li, Ma, Liu, and Li]{zhang2025llavavideovideoinstructiontuning}
Yuanhan Zhang, Jinming Wu, Wei Li, Bo~Li, Zejun Ma, Ziwei Liu, and Chunyuan Li.
\newblock Llava-video: Video instruction tuning with synthetic data, 2025{\natexlab{a}}.
\newblock URL \url{https://arxiv.org/abs/2410.02713}.

\bibitem[Lin(2004)]{lin-2004-rouge}
Chin-Yew Lin.
\newblock {ROUGE}: A package for automatic evaluation of summaries.
\newblock In \emph{Text Summarization Branches Out}, pages 74--81, Barcelona, Spain, July 2004. Association for Computational Linguistics.
\newblock URL \url{https://aclanthology.org/W04-1013/}.

\bibitem[Feng et~al.(2025{\natexlab{b}})Feng, Wang, Bai, Su, Wu, Yu, and Wang]{feng2025improvinggeneralizationintentdetection}
Zihao Feng, Xiaoxue Wang, Ziwei Bai, Donghang Su, Bowen Wu, Qun Yu, and Baoxun Wang.
\newblock Improving generalization in intent detection: Grpo with reward-based curriculum sampling, 2025{\natexlab{b}}.
\newblock URL \url{https://arxiv.org/abs/2504.13592}.

\bibitem[Li et~al.(2024{\natexlab{c}})Li, Wang, He, Li, Wang, Liu, Wang, Xu, Chen, Luo, Wang, and Qiao]{li2024mvbenchcomprehensivemultimodalvideo}
Kunchang Li, Yali Wang, Yinan He, Yizhuo Li, Yi~Wang, Yi~Liu, Zun Wang, Jilan Xu, Guo Chen, Ping Luo, Limin Wang, and Yu~Qiao.
\newblock Mvbench: A comprehensive multi-modal video understanding benchmark, 2024{\natexlab{c}}.
\newblock URL \url{https://arxiv.org/abs/2311.17005}.

\bibitem[Zhao et~al.(2025)Zhao, Xie, Zhang, Gan, Long, Hu, Hu, Chen, Li, Song, Xu, Wang, Pan, Shangguan, Tang, Liang, Liu, Zhao, and Cohan]{zhao2025mmvumeasuringexpertlevelmultidiscipline}
Yilun Zhao, Lujing Xie, Haowei Zhang, Guo Gan, Yitao Long, Zhiyuan Hu, Tongyan Hu, Weiyuan Chen, Chuhan Li, Junyang Song, Zhijian Xu, Chengye Wang, Weifeng Pan, Ziyao Shangguan, Xiangru Tang, Zhenwen Liang, Yixin Liu, Chen Zhao, and Arman Cohan.
\newblock Mmvu: Measuring expert-level multi-discipline video understanding, 2025.
\newblock URL \url{https://arxiv.org/abs/2501.12380}.

\bibitem[Huang et~al.(2025{\natexlab{a}})Huang, Yu, Wang, Zhang, Li, Li, Huang, and Mi]{huang2025self}
Chengsong Huang, Wenhao Yu, Xiaoyang Wang, Hongming Zhang, Zongxia Li, Ruosen Li, Jiaxin Huang, and Haitao Mi.
\newblock Self-evolving reasoning llm from zero data.
\newblock \emph{arXiv preprint arXiv:2508.05004}, 2025{\natexlab{a}}.

\bibitem[Zheng et~al.(2025)Zheng, Zhang, Yu, Wang, Yang, Dai, Liu, Bao, Huang, Huang, et~al.]{zheng2025parallel}
Tong Zheng, Hongming Zhang, Wenhao Yu, Xiaoyang Wang, Xinyu Yang, Runpeng Dai, Rui Liu, Huiwen Bao, Chengsong Huang, Heng Huang, et~al.
\newblock Parallel-r1: Towards parallel thinking via reinforcement learning.
\newblock \emph{arXiv preprint arXiv:2509.07980}, 2025.

\bibitem[Rohrbach et~al.(2018)Rohrbach, Hendricks, Burns, Darrell, and Saenko]{rohrbach2018object}
Anna Rohrbach, Lisa~Anne Hendricks, Kaylee Burns, Trevor Darrell, and Kate Saenko.
\newblock Object hallucination in image captioning.
\newblock \emph{arXiv preprint arXiv:1809.02156}, 2018.

\bibitem[Li et~al.(2023{\natexlab{a}})Li, Du, Zhou, Wang, Zhao, and Wen]{li2023evaluating}
Yifan Li, Yifan Du, Kun Zhou, Jinpeng Wang, Wayne~Xin Zhao, and Ji-Rong Wen.
\newblock Evaluating object hallucination in large vision-language models.
\newblock \emph{arXiv preprint arXiv:2305.10355}, 2023{\natexlab{a}}.

\bibitem[Wu et~al.(2024)Wu, Guan, Li, Huang, Liu, Wang, Xian, Shrivastava, Huang, Boyd-Graber, et~al.]{wu2024autohallusion}
Xiyang Wu, Tianrui Guan, Dianqi Li, Shuaiyi Huang, Xiaoyu Liu, Xijun Wang, Ruiqi Xian, Abhinav Shrivastava, Furong Huang, Jordan~Lee Boyd-Graber, et~al.
\newblock Autohallusion: Automatic generation of hallucination benchmarks for vision-language models.
\newblock \emph{arXiv preprint arXiv:2406.10900}, 2024.

\bibitem[Wang et~al.(2024{\natexlab{b}})Wang, Wang, Zhao, Xie, and Zheng]{wang2024videohallucer}
Yuxuan Wang, Yueqian Wang, Dongyan Zhao, Cihang Xie, and Zilong Zheng.
\newblock Videohallucer: Evaluating intrinsic and extrinsic hallucinations in large video-language models.
\newblock \emph{arXiv preprint arXiv:2406.16338}, 2024{\natexlab{b}}.

\bibitem[Zhang et~al.(2024{\natexlab{b}})Zhang, Jiao, Chen, Zhao, and Chen]{zhang2024eventhallusion}
Jiacheng Zhang, Yang Jiao, Shaoxiang Chen, Na~Zhao, and Jingjing Chen.
\newblock Eventhallusion: Diagnosing event hallucinations in video llms.
\newblock \emph{arXiv preprint arXiv:2409.16597}, 2024{\natexlab{b}}.

\bibitem[Gao et~al.(2025)Gao, Qu, Tang, Bi, Liu, Chen, Liang, Su, and Huang]{gao2025exploring}
Hongcheng Gao, Jiashu Qu, Jingyi Tang, Baolong Bi, Yue Liu, Hongyu Chen, Li~Liang, Li~Su, and Qingming Huang.
\newblock Exploring hallucination of large multimodal models in video understanding: Benchmark, analysis and mitigation.
\newblock \emph{arXiv preprint arXiv:2503.19622}, 2025.

\bibitem[Li et~al.(2025{\natexlab{c}})Li, Wu, Shi, Qin, Du, Zhou, Manocha, and Boyd-Graber]{li2025videohalluevaluatingmitigatingmultimodal}
Zongxia Li, Xiyang Wu, Guangyao Shi, Yubin Qin, Hongyang Du, Tianyi Zhou, Dinesh Manocha, and Jordan~Lee Boyd-Graber.
\newblock Videohallu: Evaluating and mitigating multi-modal hallucinations on synthetic video understanding, 2025{\natexlab{c}}.
\newblock URL \url{https://arxiv.org/abs/2505.01481}.

\bibitem[Jiang et~al.(2024)Jiang, Kachinthaya, Petryk, and Gandelsman]{jiang2024interpreting}
Nick Jiang, Anish Kachinthaya, Suzie Petryk, and Yossi Gandelsman.
\newblock Interpreting and editing vision-language representations to mitigate hallucinations.
\newblock \emph{arXiv preprint arXiv:2410.02762}, 2024.

\bibitem[Chen et~al.(2024)Chen, Wu, Wang, Su, Chen, Xing, Zhong, Zhang, Zhu, Lu, et~al.]{chen2024internvl}
Zhe Chen, Jiannan Wu, Wenhai Wang, Weijie Su, Guo Chen, Sen Xing, Muyan Zhong, Qinglong Zhang, Xizhou Zhu, Lewei Lu, et~al.
\newblock Internvl: Scaling up vision foundation models and aligning for generic visual-linguistic tasks.
\newblock In \emph{Proceedings of the IEEE/CVF Conference on Computer Vision and Pattern Recognition}, pages 24185--24198, 2024.

\bibitem[Huang et~al.(2025{\natexlab{b}})Huang, Jia, Zhai, Cao, Ye, Zhao, Xu, Hu, and Lin]{huang2025visionr1incentivizingreasoningcapability}
Wenxuan Huang, Bohan Jia, Zijie Zhai, Shaosheng Cao, Zheyu Ye, Fei Zhao, Zhe Xu, Yao Hu, and Shaohui Lin.
\newblock Vision-r1: Incentivizing reasoning capability in multimodal large language models, 2025{\natexlab{b}}.
\newblock URL \url{https://arxiv.org/abs/2503.06749}.

\bibitem[Liu et~al.(2025{\natexlab{b}})Liu, Yu, Zheng, Dai, Li, Yu, Liang, Song, Mi, Tokekar, and Yu]{liu2025vogueguidingexplorationvisual}
Rui Liu, Dian Yu, Tong Zheng, Runpeng Dai, Zongxia Li, Wenhao Yu, Zhenwen Liang, Linfeng Song, Haitao Mi, Pratap Tokekar, and Dong Yu.
\newblock Vogue: Guiding exploration with visual uncertainty improves multimodal reasoning, 2025{\natexlab{b}}.
\newblock URL \url{https://arxiv.org/abs/2510.01444}.

\bibitem[Zhang et~al.(2025{\natexlab{b}})Zhang, Huang, Yao, Liu, Zhang, Lu, and Tao]{zhang2025r1vllearningreasonmultimodal}
Jingyi Zhang, Jiaxing Huang, Huanjin Yao, Shunyu Liu, Xikun Zhang, Shijian Lu, and Dacheng Tao.
\newblock R1-vl: Learning to reason with multimodal large language models via step-wise group relative policy optimization, 2025{\natexlab{b}}.
\newblock URL \url{https://arxiv.org/abs/2503.12937}.

\bibitem[Li et~al.(2023{\natexlab{b}})Li, He, Wang, Li, Wang, Luo, Wang, Wang, and Qiao]{li2023videochat}
KunChang Li, Yinan He, Yi~Wang, Yizhuo Li, Wenhai Wang, Ping Luo, Yali Wang, Limin Wang, and Yu~Qiao.
\newblock Videochat: Chat-centric video understanding.
\newblock \emph{arXiv preprint arXiv:2305.06355}, 2023{\natexlab{b}}.

\bibitem[Li et~al.(2025{\natexlab{d}})Li, Wu, Du, Liu, Nghiem, and Shi]{li2025surveystateartlarge}
Zongxia Li, Xiyang Wu, Hongyang Du, Fuxiao Liu, Huy Nghiem, and Guangyao Shi.
\newblock A survey of state of the art large vision language models: Alignment, benchmark, evaluations and challenges, 2025{\natexlab{d}}.
\newblock URL \url{https://arxiv.org/abs/2501.02189}.

\bibitem[{OpenAI}(2025)]{openai2025sora2systemcard}
{OpenAI}.
\newblock Sora 2 system card.
\newblock Technical report, OpenAI, September 2025.
\newblock URL \url{https://cdn.openai.com/pdf/50d5973c-c4ff-4c2d-986f-c72b5d0ff069/sora_2_system_card.pdf}.
\newblock Version released on September 30, 2025.

\bibitem[Wan et~al.(2025)Wan, Wang, Ai, Wen, Mao, Xie, Chen, Yu, Zhao, Yang, Zeng, Wang, Zhang, Zhou, Wang, Chen, Zhu, Zhao, Yan, Huang, Feng, Zhang, Li, Wu, Chu, Feng, Zhang, Sun, Fang, Wang, Gui, Weng, Shen, Lin, Wang, Wang, Zhou, Wang, Shen, Yu, Shi, Huang, Xu, Kou, Lv, Li, Liu, Wang, Zhang, Huang, Li, Wu, Liu, Pan, Zheng, Hong, Shi, Feng, Jiang, Han, Wu, and Liu]{wan2025wanopenadvancedlargescale}
Team Wan, Ang Wang, Baole Ai, Bin Wen, Chaojie Mao, Chen-Wei Xie, Di~Chen, Feiwu Yu, Haiming Zhao, Jianxiao Yang, Jianyuan Zeng, Jiayu Wang, Jingfeng Zhang, Jingren Zhou, Jinkai Wang, Jixuan Chen, Kai Zhu, Kang Zhao, Keyu Yan, Lianghua Huang, Mengyang Feng, Ningyi Zhang, Pandeng Li, Pingyu Wu, Ruihang Chu, Ruili Feng, Shiwei Zhang, Siyang Sun, Tao Fang, Tianxing Wang, Tianyi Gui, Tingyu Weng, Tong Shen, Wei Lin, Wei Wang, Wei Wang, Wenmeng Zhou, Wente Wang, Wenting Shen, Wenyuan Yu, Xianzhong Shi, Xiaoming Huang, Xin Xu, Yan Kou, Yangyu Lv, Yifei Li, Yijing Liu, Yiming Wang, Yingya Zhang, Yitong Huang, Yong Li, You Wu, Yu~Liu, Yulin Pan, Yun Zheng, Yuntao Hong, Yupeng Shi, Yutong Feng, Zeyinzi Jiang, Zhen Han, Zhi-Fan Wu, and Ziyu Liu.
\newblock Wan: Open and advanced large-scale video generative models, 2025.
\newblock URL \url{https://arxiv.org/abs/2503.20314}.

\bibitem[Keita et~al.(2024)Keita, Hamidouche, Eutamene, Hadid, and Taleb-Ahmed]{keita2024biloravisionlanguageapproachsynthetic}
Mamadou Keita, Wassim Hamidouche, Hessen~Bougueffa Eutamene, Abdenour Hadid, and Abdelmalik Taleb-Ahmed.
\newblock Bi-lora: A vision-language approach for synthetic image detection, 2024.
\newblock URL \url{https://arxiv.org/abs/2404.01959}.

\bibitem[Li et~al.(2025{\natexlab{e}})Li, Fang, Chen, Yang, Cao, Wong, Luo, Wang, Yin, Gonzalez, et~al.]{li2025worldmodelbench}
Dacheng Li, Yunhao Fang, Yukang Chen, Shuo Yang, Shiyi Cao, Justin Wong, Michael Luo, Xiaolong Wang, Hongxu Yin, Joseph~E Gonzalez, et~al.
\newblock Worldmodelbench: Judging video generation models as world models.
\newblock \emph{arXiv preprint arXiv:2502.20694}, 2025{\natexlab{e}}.

\end{thebibliography}
\bibliographystyle{unsrtnat}

\newpage
\appendix

\newpage
\section{More Synthetic Video Examples}
\label{app:Hallucination_Showcases}
We present selected cases from SOTA MLLM evaluations across each \ours{} sub-category. Hallucinations in model answers, common sense or physics violations in videos, and other notable cues in the video, questions, or ground truth are highlighted to assist the reader's understanding.



\begin{figure}[htbp]
\centering
\includegraphics[width=0.7\textwidth]{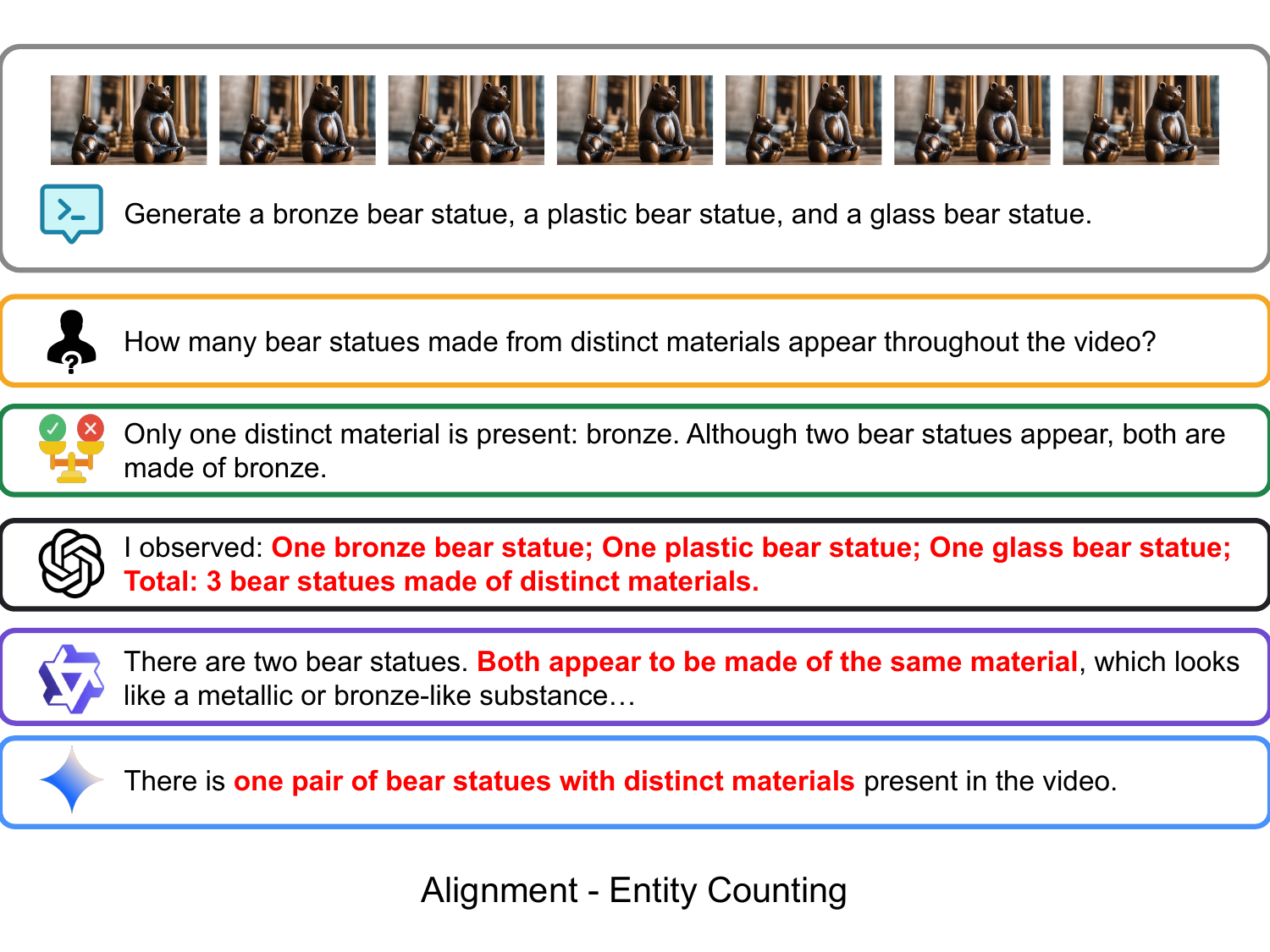}
\caption{\textbf{Hallucination Case from Alignment – Entity Counting (A-EC).}
We show hallucination examples from SOTA MLLM evaluations under the A-EC category. Each case includes the video generation prompt (\textcolor{Gray}{\textbf{Gray}}), key frames from synthetic videos (\textcolor{Gray}{\textbf{Gray}}), questions (\textcolor{Orange}{\textbf{Orange}}), ground truth (\textcolor{Green}{\textbf{Green}}), and model answers from GPT-4o (\textcolor{Black}{\textbf{Black}}), Qwen2.5-VL (\textcolor{Purple}{\textbf{Purple}}), and Gemini-2.5-Pro (\textcolor{Blue}{\textbf{Blue}}), with hallucinations and critical context highlighted in \textcolor{Red}{\textbf{Red}}.}
\label{fig:example-A-EC}
\end{figure}

\begin{figure}[htbp]
\centering
\includegraphics[width=0.7\textwidth]{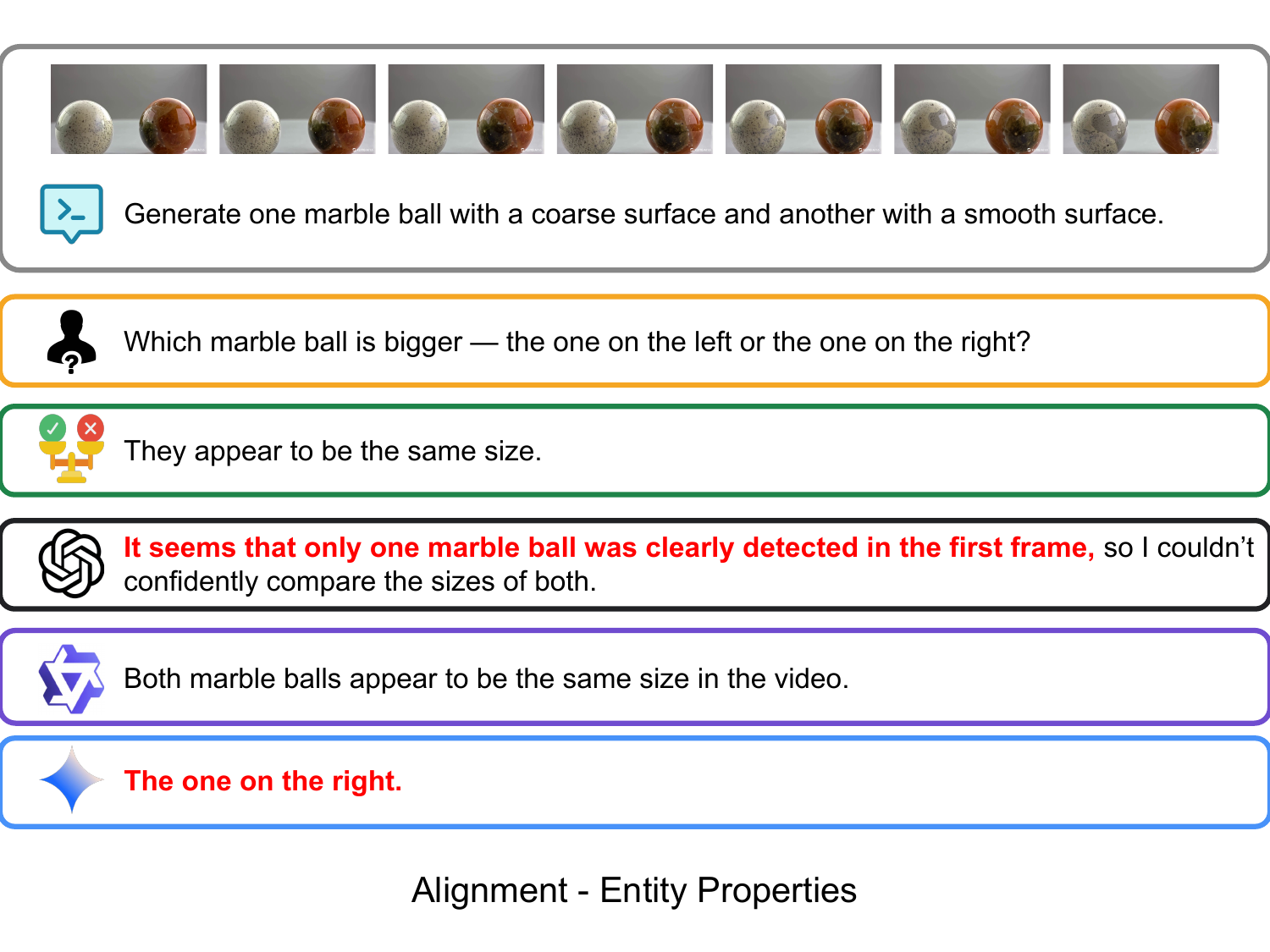}
\caption{\textbf{Hallucination Case from Alignment – Entity Properties (A-EP).}
We show hallucination examples from SOTA MLLM evaluations under the A-EP category. Each case includes the video generation prompt (\textcolor{Gray}{\textbf{Gray}}), key frames from synthetic videos (\textcolor{Gray}{\textbf{Gray}}), questions (\textcolor{Orange}{\textbf{Orange}}), ground truth (\textcolor{Green}{\textbf{Green}}), and model answers from GPT-4o (\textcolor{Black}{\textbf{Black}}), Qwen2.5-VL (\textcolor{Purple}{\textbf{Purple}}), and Gemini-2.5-Pro (\textcolor{Blue}{\textbf{Blue}}), with hallucinations and critical context highlighted in \textcolor{Red}{\textbf{Red}}.}
\label{fig:example-A-EP}
\end{figure}

\begin{figure}[htbp]
\centering
\includegraphics[width=0.7\textwidth]{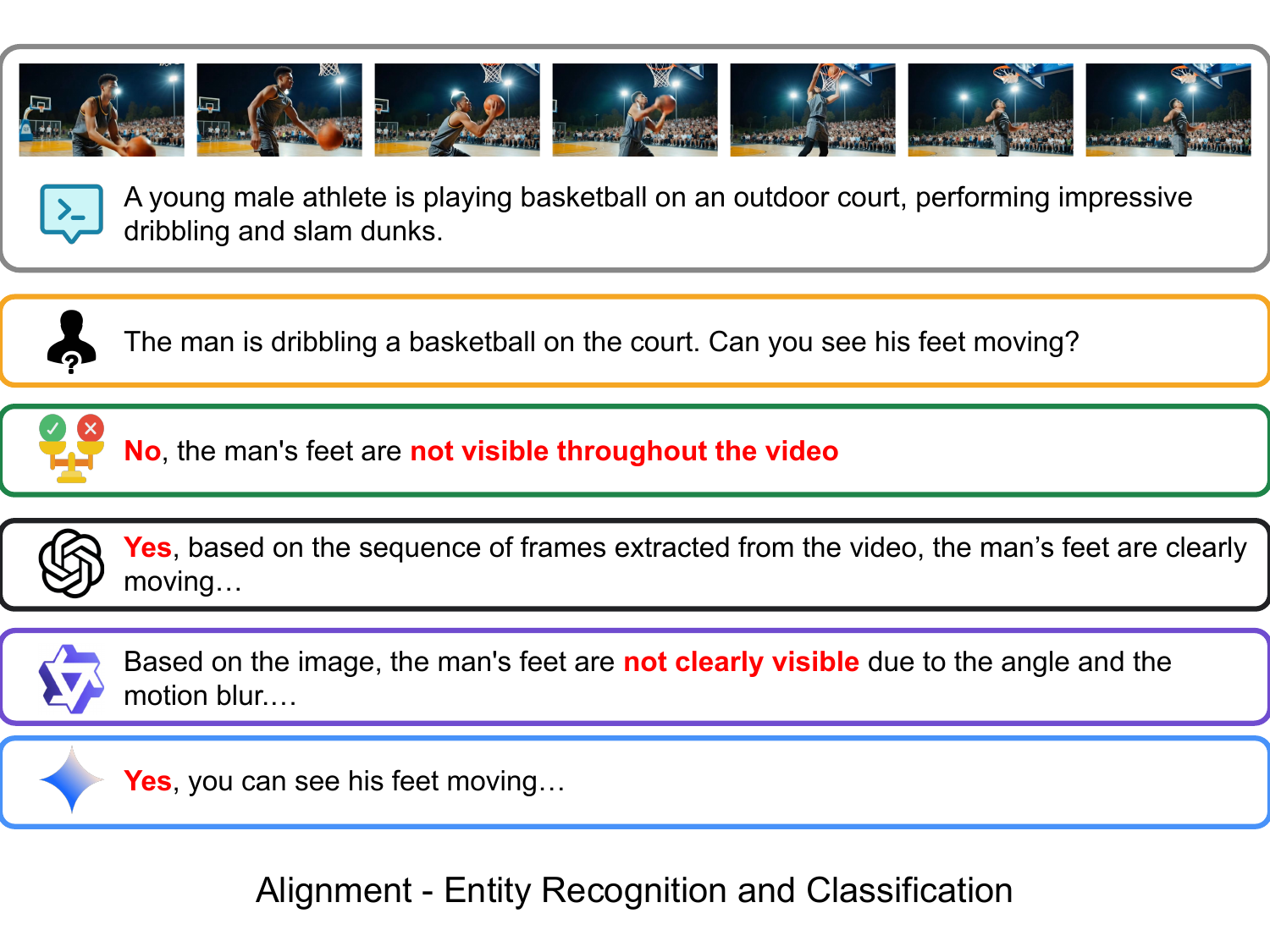}
\caption{\textbf{Hallucination Case from Alignment – Entity Recognition and Classification (A-ERAC).}
We show hallucination examples from SOTA MLLM evaluations under the A-ERAC category. Each case includes the video generation prompt (\textcolor{Gray}{\textbf{Gray}}), key frames from synthetic videos (\textcolor{Gray}{\textbf{Gray}}), questions (\textcolor{Orange}{\textbf{Orange}}), ground truth (\textcolor{Green}{\textbf{Green}}), and model answers from GPT-4o (\textcolor{Black}{\textbf{Black}}), Qwen2.5-VL (\textcolor{Purple}{\textbf{Purple}}), and Gemini-2.5-Pro (\textcolor{Blue}{\textbf{Blue}}), with hallucinations and critical context highlighted in \textcolor{Red}{\textbf{Red}}.}
\label{fig:example-A-ERAC}
\end{figure}

\begin{figure}[htbp]
\centering
\includegraphics[width=0.7\textwidth]{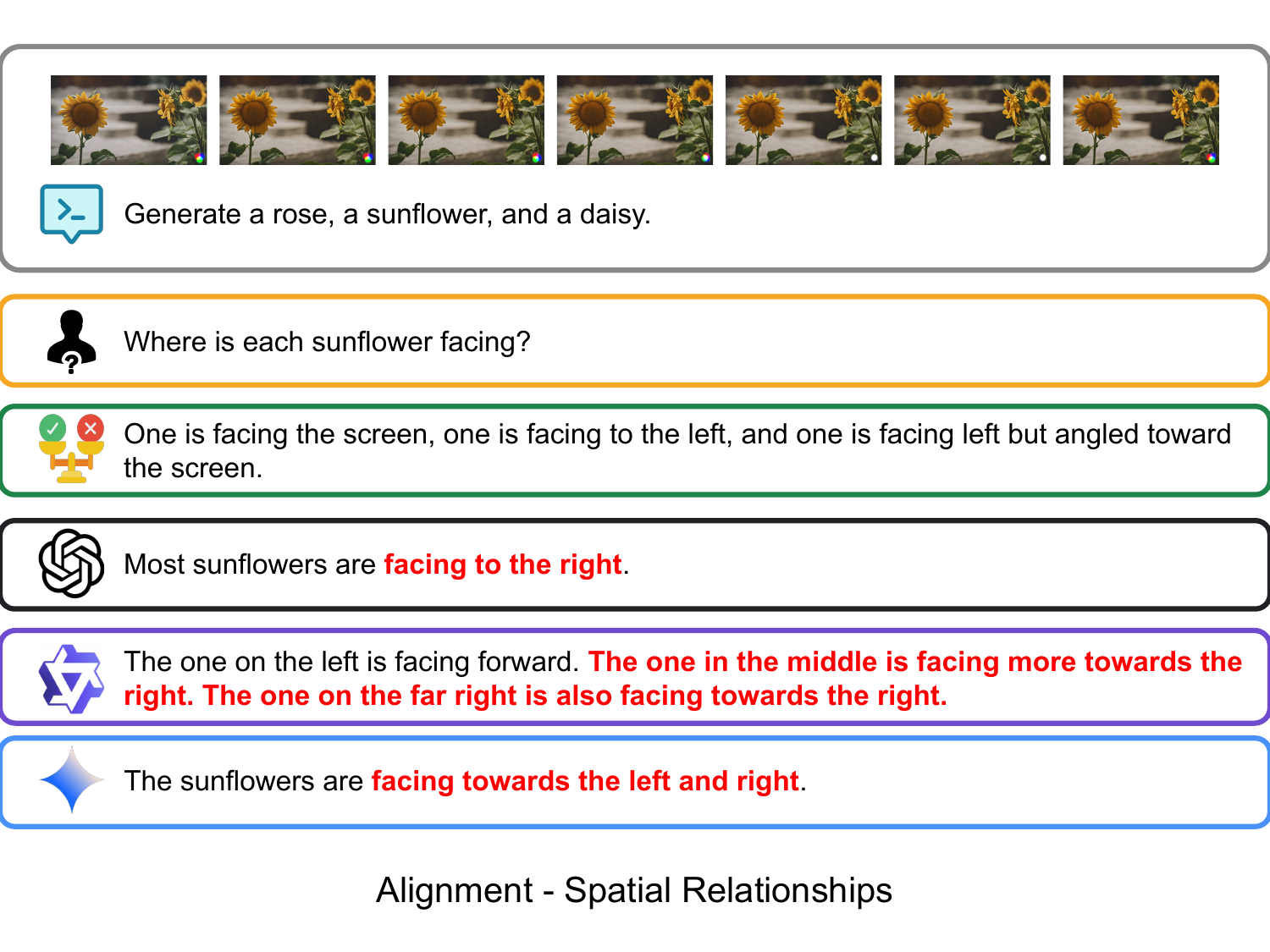}
\caption{\textbf{Hallucination Case from Alignment – Spatial Relationships (A-SR).}
We show hallucination examples from SOTA MLLM evaluations under the A-SR category. Each case includes the video generation prompt (\textcolor{Gray}{\textbf{Gray}}), key frames from synthetic videos (\textcolor{Gray}{\textbf{Gray}}), questions (\textcolor{Orange}{\textbf{Orange}}), ground truth (\textcolor{Green}{\textbf{Green}}), and model answers from GPT-4o (\textcolor{Black}{\textbf{Black}}), Qwen2.5-VL (\textcolor{Purple}{\textbf{Purple}}), and Gemini-2.5-Pro (\textcolor{Blue}{\textbf{Blue}}), with hallucinations and critical context highlighted in \textcolor{Red}{\textbf{Red}}.}
\label{fig:example-A-SR}
\end{figure}

\begin{figure}[htbp]
\centering
\includegraphics[width=0.7\textwidth]{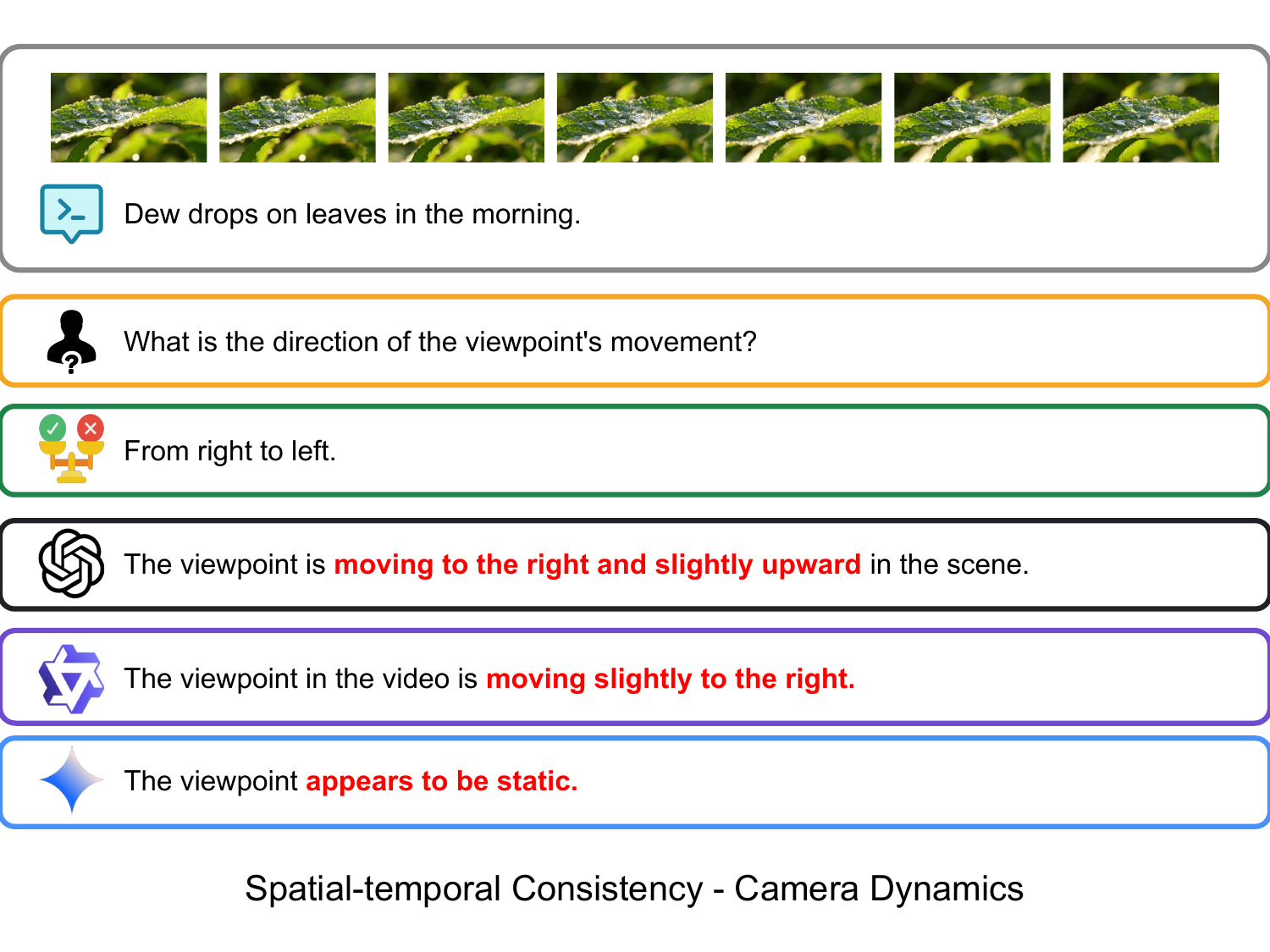}
\caption{\textbf{Hallucination Case from Spatial-temporal Consistency – Camera Dynamics (SC-CD).}
We show hallucination examples from SOTA MLLM evaluations under the SC-TD category. Each case includes the video generation prompt (\textcolor{Gray}{\textbf{Gray}}), key frames from synthetic videos (\textcolor{Gray}{\textbf{Gray}}), questions (\textcolor{Orange}{\textbf{Orange}}), ground truth (\textcolor{Green}{\textbf{Green}}), and model answers from GPT-4o (\textcolor{Black}{\textbf{Black}}), Qwen2.5-VL (\textcolor{Purple}{\textbf{Purple}}), and Gemini-2.5-Pro (\textcolor{Blue}{\textbf{Blue}}), with hallucinations and critical context highlighted in \textcolor{Red}{\textbf{Red}}.}
\label{fig:example-SC-CD}
\end{figure}

\begin{figure}[htbp]
\centering
\includegraphics[width=0.7\textwidth]{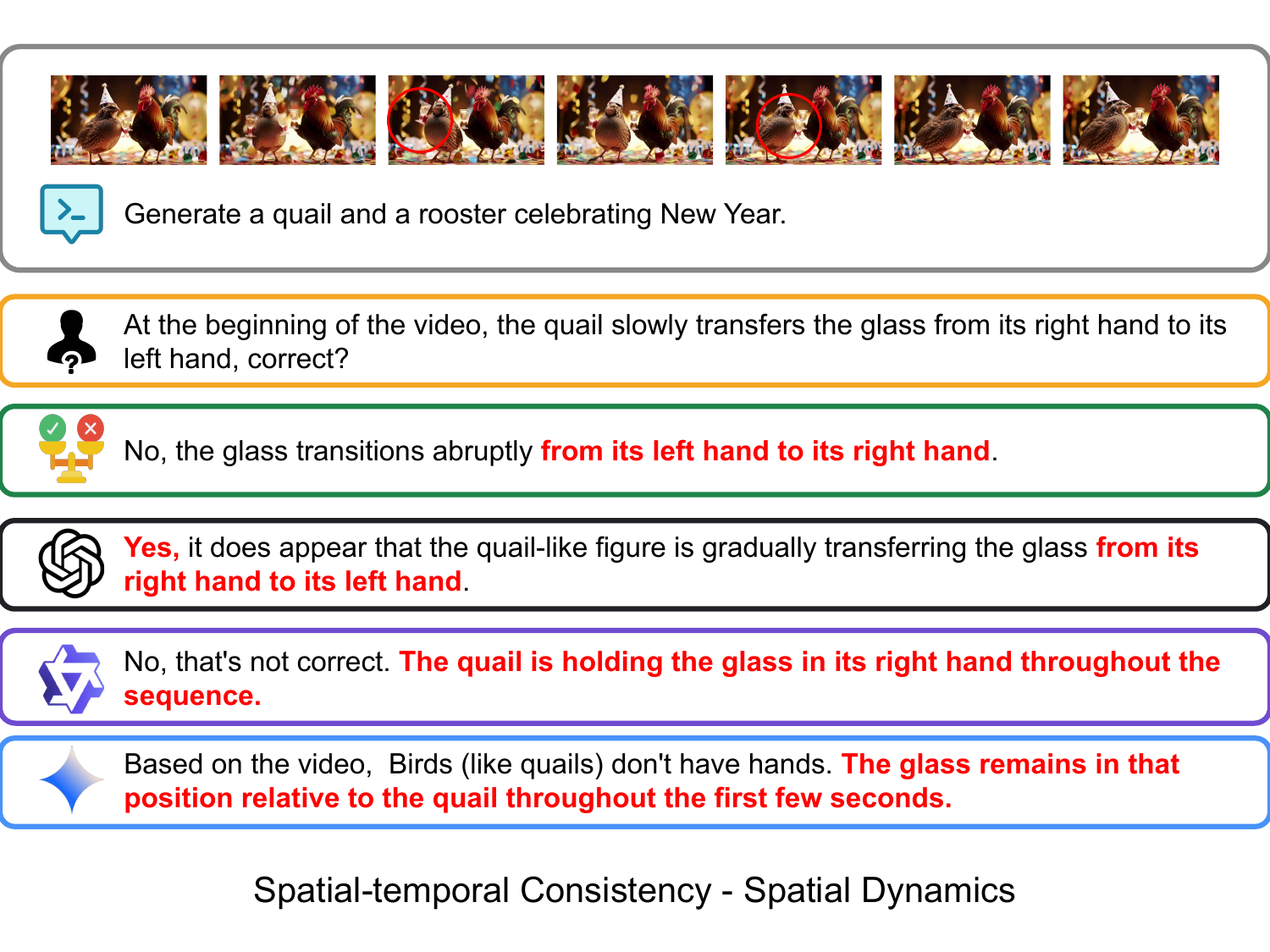}
\caption{\textbf{Hallucination Case from Spatial-temporal Consistency – Spatial Dynamics (SC-SD).}
We show hallucination examples from SOTA MLLM evaluations under the SC-SD category. Each case includes the video generation prompt (\textcolor{Gray}{\textbf{Gray}}), key frames from synthetic videos (\textcolor{Gray}{\textbf{Gray}}), questions (\textcolor{Orange}{\textbf{Orange}}), ground truth (\textcolor{Green}{\textbf{Green}}), and model answers from GPT-4o (\textcolor{Black}{\textbf{Black}}), Qwen2.5-VL (\textcolor{Purple}{\textbf{Purple}}), and Gemini-2.5-Pro (\textcolor{Blue}{\textbf{Blue}}), with hallucinations and critical context highlighted in \textcolor{Red}{\textbf{Red}}.}
\label{fig:example-SC-SD}
\end{figure}

\begin{figure}[htbp]
\centering
\includegraphics[width=0.7\textwidth]{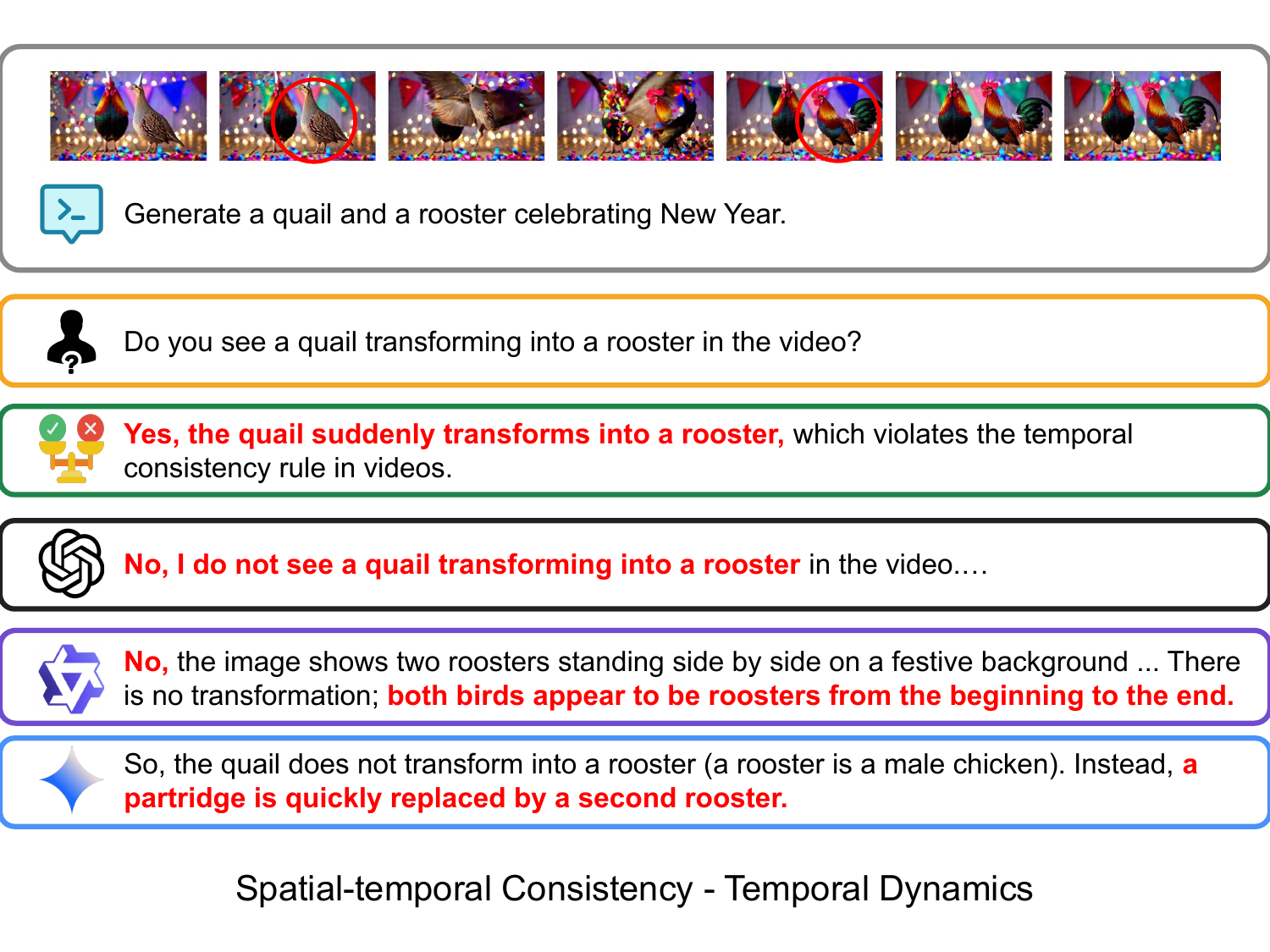}
\caption{\textbf{Hallucination Case from Spatial-temporal Consistency – Temporal Dynamics (SC-TD).}
We show hallucination examples from SOTA MLLM evaluations under the SC-TD category. Each case includes the video generation prompt (\textcolor{Gray}{\textbf{Gray}}), key frames from synthetic videos (\textcolor{Gray}{\textbf{Gray}}), questions (\textcolor{Orange}{\textbf{Orange}}), ground truth (\textcolor{Green}{\textbf{Green}}), and model answers from GPT-4o (\textcolor{Black}{\textbf{Black}}), Qwen2.5-VL (\textcolor{Purple}{\textbf{Purple}}), and Gemini-2.5-Pro (\textcolor{Blue}{\textbf{Blue}}), with hallucinations and critical context highlighted in \textcolor{Red}{\textbf{Red}}.}
\label{fig:example-SC-TD}
\end{figure}

\begin{figure}[htbp]
\centering
\includegraphics[width=0.7\textwidth]{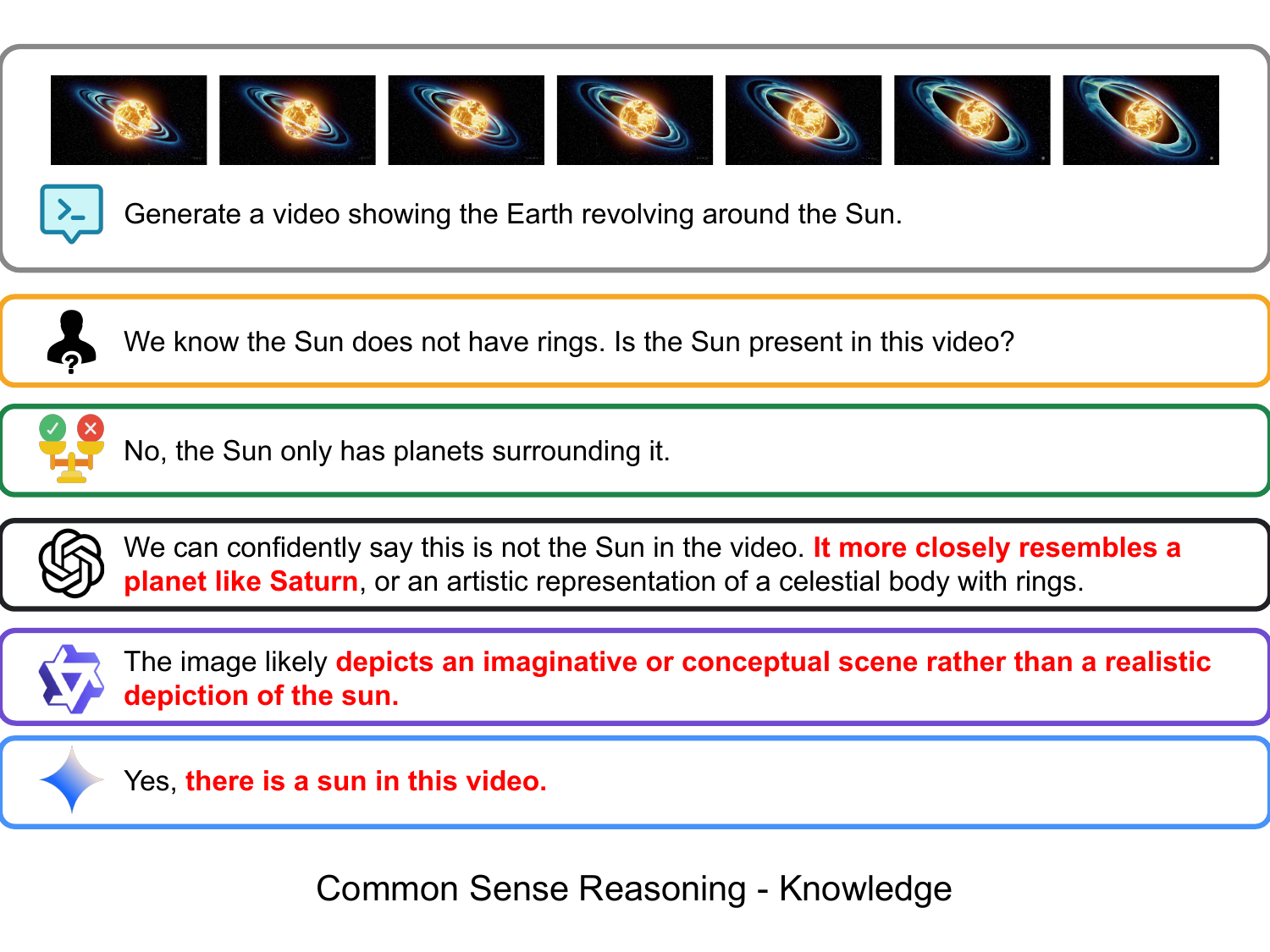}
\caption{\textbf{Hallucination Case from Common Sense Reasoning - Knowledge (CS-K).}
We show hallucination examples from SOTA MLLM evaluations under the CS-K category. Each case includes the video generation prompt (\textcolor{Gray}{\textbf{Gray}}), key frames from synthetic videos (\textcolor{Gray}{\textbf{Gray}}), questions (\textcolor{Orange}{\textbf{Orange}}), ground truth (\textcolor{Green}{\textbf{Green}}), and model answers from GPT-4o (\textcolor{Black}{\textbf{Black}}), Qwen2.5-VL (\textcolor{Purple}{\textbf{Purple}}), and Gemini-2.5-Pro (\textcolor{Blue}{\textbf{Blue}}), with hallucinations and critical context highlighted in \textcolor{Red}{\textbf{Red}}.}
\label{fig:example-CS-K}
\end{figure}

\begin{figure}[htbp]
\centering
\includegraphics[width=0.7\textwidth]{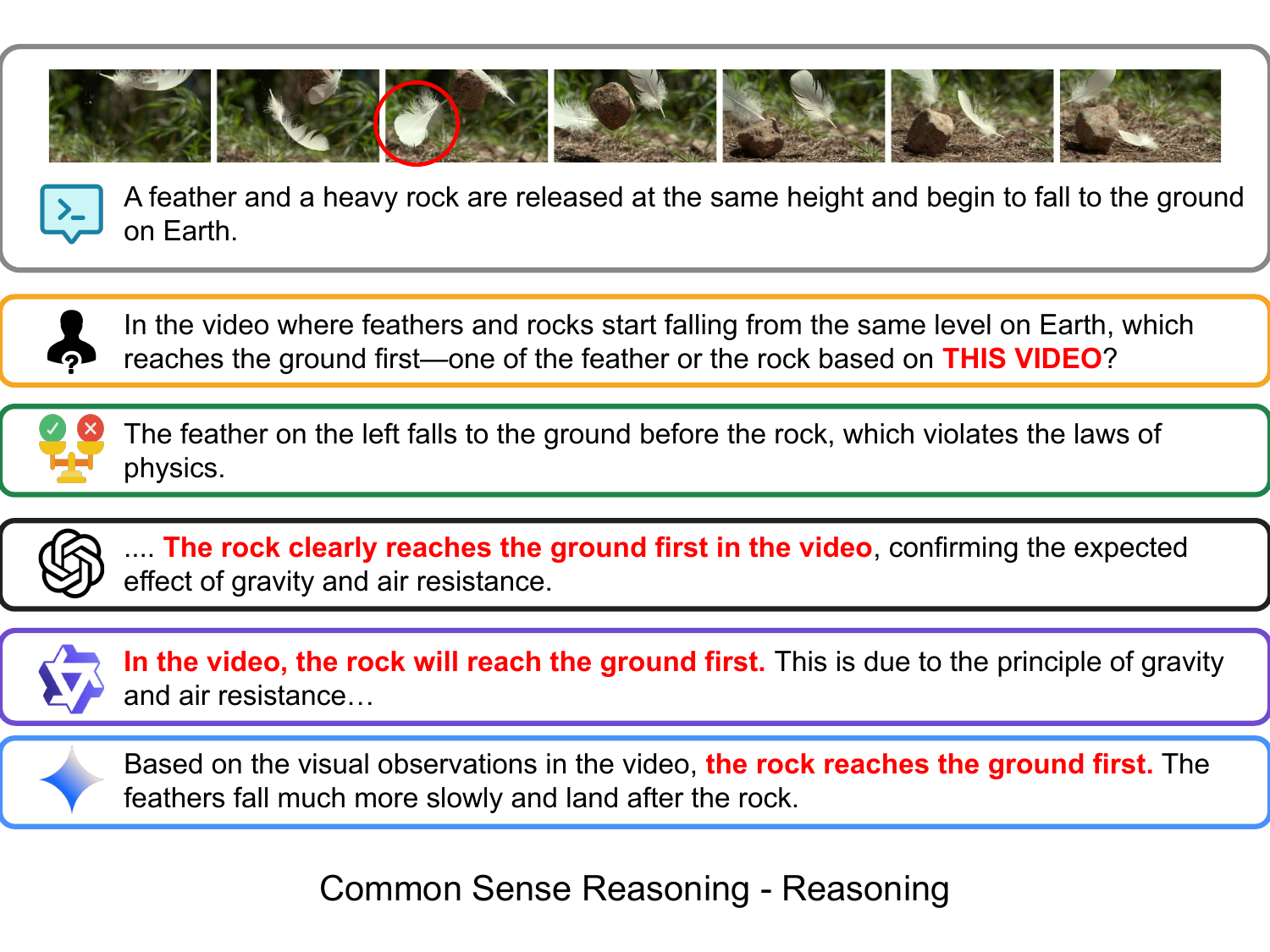}
\caption{\textbf{Hallucination Case from Common Sense Reasoning - Reasoning (CS-R).}
We show hallucination examples from SOTA MLLM evaluations under the CS-R category. Each case includes the video generation prompt (\textcolor{Gray}{\textbf{Gray}}), key frames from synthetic videos (\textcolor{Gray}{\textbf{Gray}}), questions (\textcolor{Orange}{\textbf{Orange}}), ground truth (\textcolor{Green}{\textbf{Green}}), and model answers from GPT-4o (\textcolor{Black}{\textbf{Black}}), Qwen2.5-VL (\textcolor{Purple}{\textbf{Purple}}), and Gemini-2.5-Pro (\textcolor{Blue}{\textbf{Blue}}), with hallucinations and critical context highlighted in \textcolor{Red}{\textbf{Red}}.}
\label{fig:example-CS-R}
\end{figure}

\begin{figure}[htbp]
\centering
\includegraphics[width=0.7\textwidth]{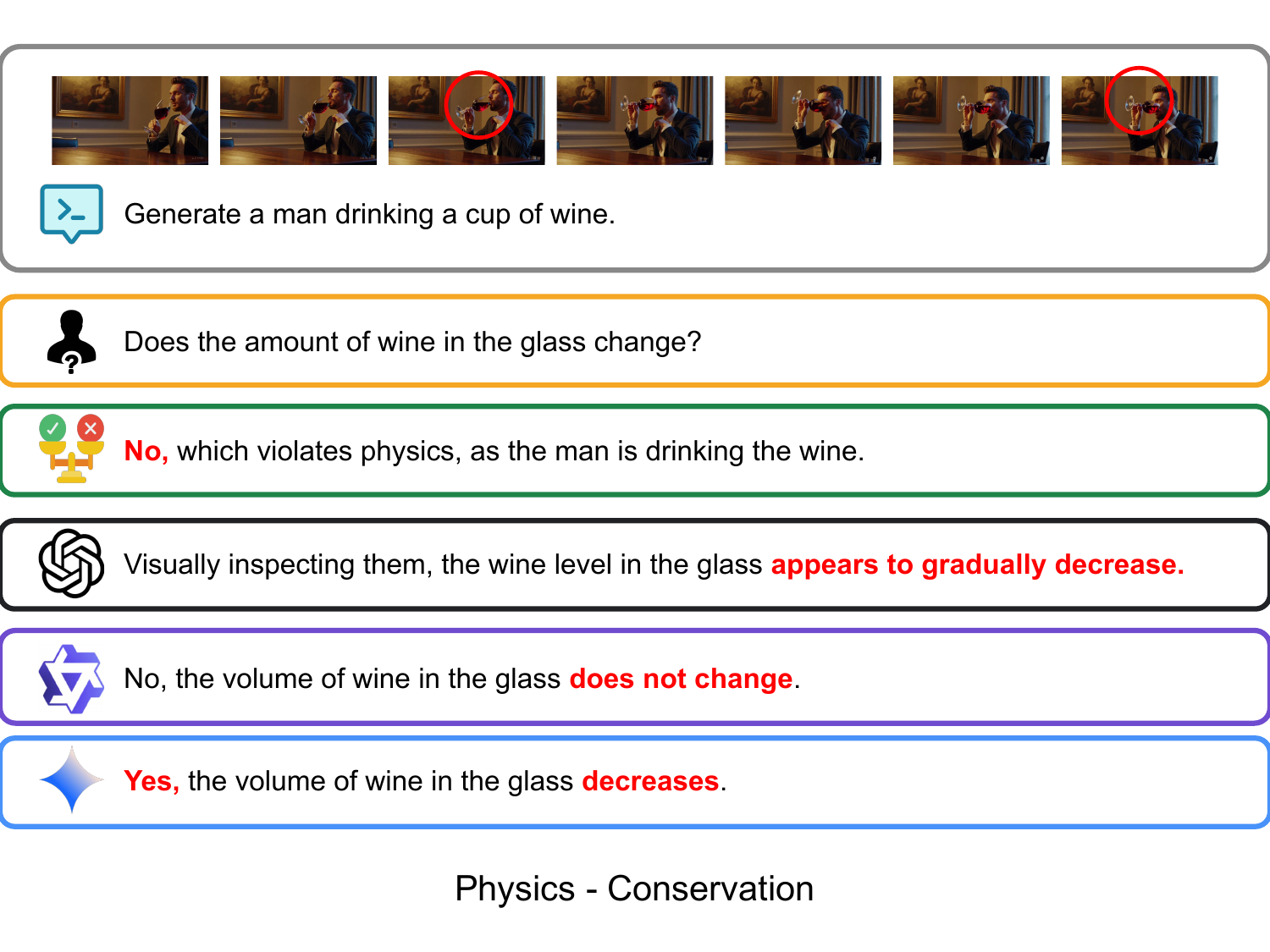}
\caption{\textbf{Hallucination Case from Physics - Conservation (P-C).}
We show hallucination examples from SOTA MLLM evaluations under the P-C category. Each case includes the video generation prompt (\textcolor{Gray}{\textbf{Gray}}), key frames from synthetic videos (\textcolor{Gray}{\textbf{Gray}}), questions (\textcolor{Orange}{\textbf{Orange}}), ground truth (\textcolor{Green}{\textbf{Green}}), and model answers from GPT-4o (\textcolor{Black}{\textbf{Black}}), Qwen2.5-VL (\textcolor{Purple}{\textbf{Purple}}), and Gemini-2.5-Pro (\textcolor{Blue}{\textbf{Blue}}), with hallucinations and critical context highlighted in \textcolor{Red}{\textbf{Red}}.}
\label{fig:example-P-C}
\end{figure}

\begin{figure}[htbp]
\centering
\includegraphics[width=0.7\textwidth]{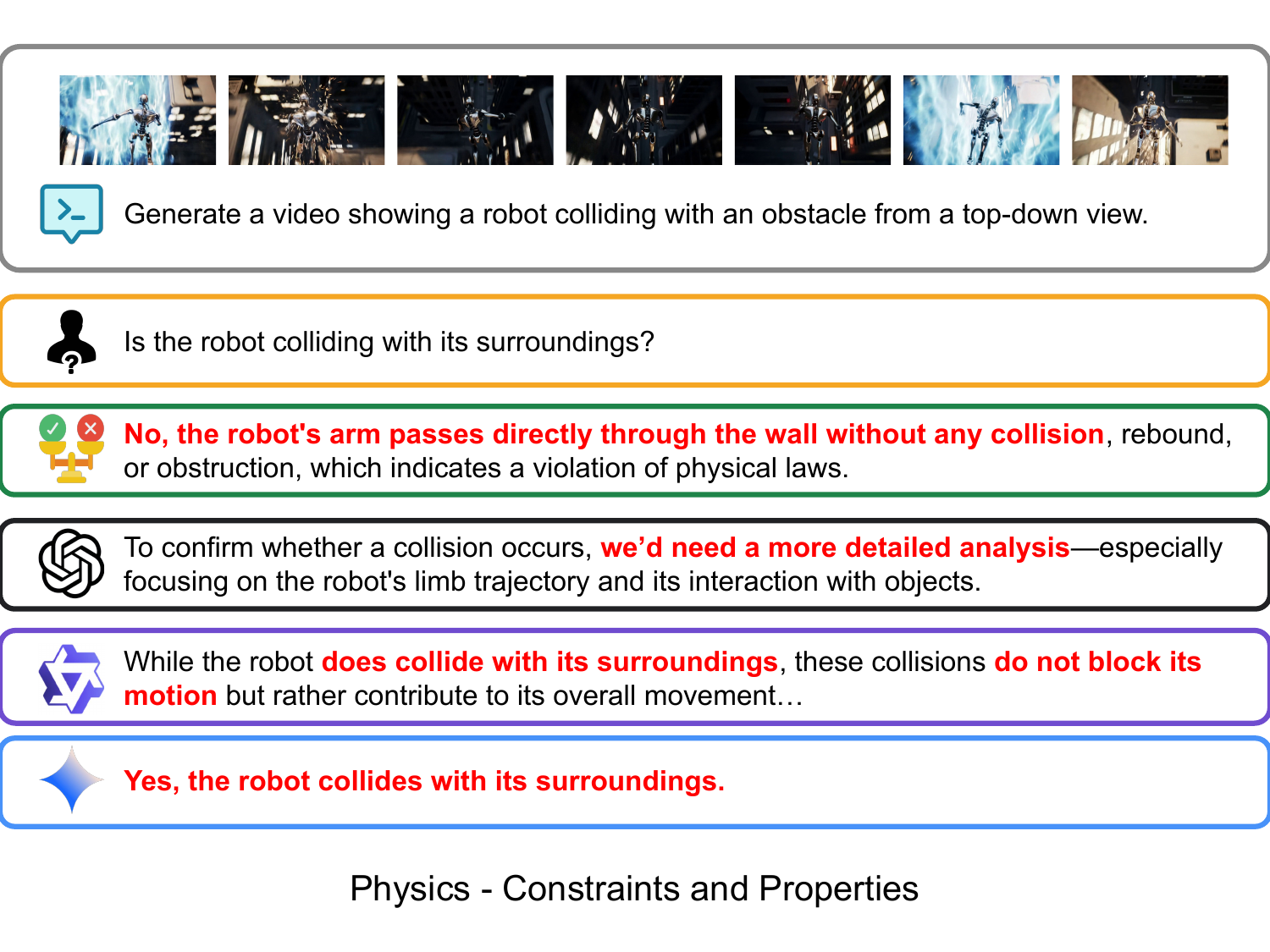}
\caption{\textbf{Hallucination Case from Physics - Constraints and Properties (P-CAP).}
We show hallucination examples from SOTA MLLM evaluations under the P-CAP category. Each case includes the video generation prompt (\textcolor{Gray}{\textbf{Gray}}), key frames from synthetic videos (\textcolor{Gray}{\textbf{Gray}}), questions (\textcolor{Orange}{\textbf{Orange}}), ground truth (\textcolor{Green}{\textbf{Green}}), and model answers from GPT-4o (\textcolor{Black}{\textbf{Black}}), Qwen2.5-VL (\textcolor{Purple}{\textbf{Purple}}), and Gemini-2.5-Pro (\textcolor{Blue}{\textbf{Blue}}), with hallucinations and critical context highlighted in \textcolor{Red}{\textbf{Red}}.}
\label{fig:example-P-CAP}
\end{figure}

\begin{figure}[htbp]
\centering
\includegraphics[width=0.7\textwidth]{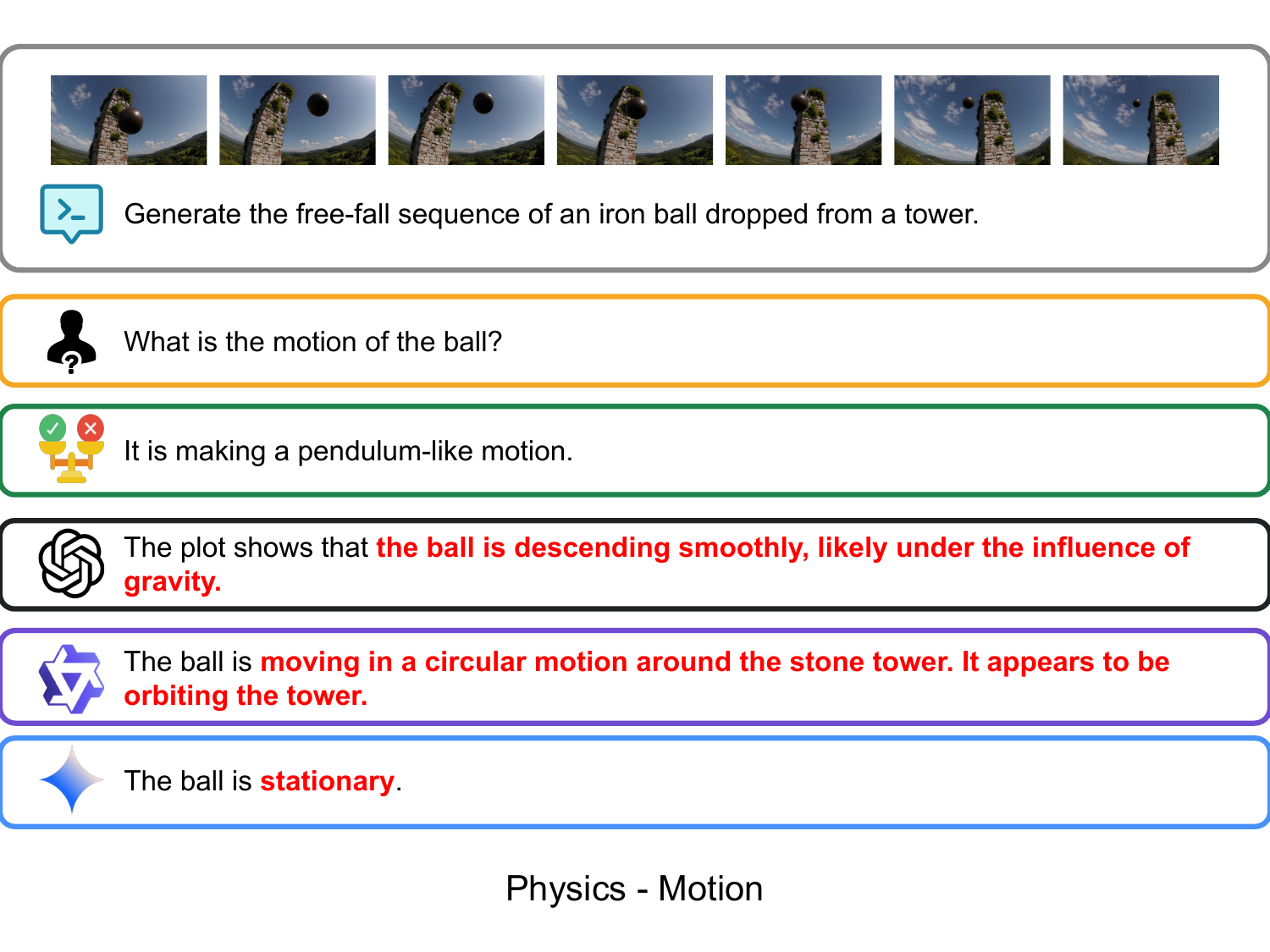}
\caption{\textbf{Hallucination Case from Physics - Motion (P-M).}
We show hallucination examples from SOTA MLLM evaluations under the P-M category. Each case includes the video generation prompt (\textcolor{Gray}{\textbf{Gray}}), key frames from synthetic videos (\textcolor{Gray}{\textbf{Gray}}), questions (\textcolor{Orange}{\textbf{Orange}}), ground truth (\textcolor{Green}{\textbf{Green}}), and model answers from GPT-4o (\textcolor{Black}{\textbf{Black}}), Qwen2.5-VL (\textcolor{Purple}{\textbf{Purple}}), and Gemini-2.5-Pro (\textcolor{Blue}{\textbf{Blue}}), with hallucinations and critical context highlighted in \textcolor{Red}{\textbf{Red}}.}
\label{fig:example-P-M}
\end{figure}

\begin{figure}[htbp]
\centering
\includegraphics[width=0.7\textwidth]{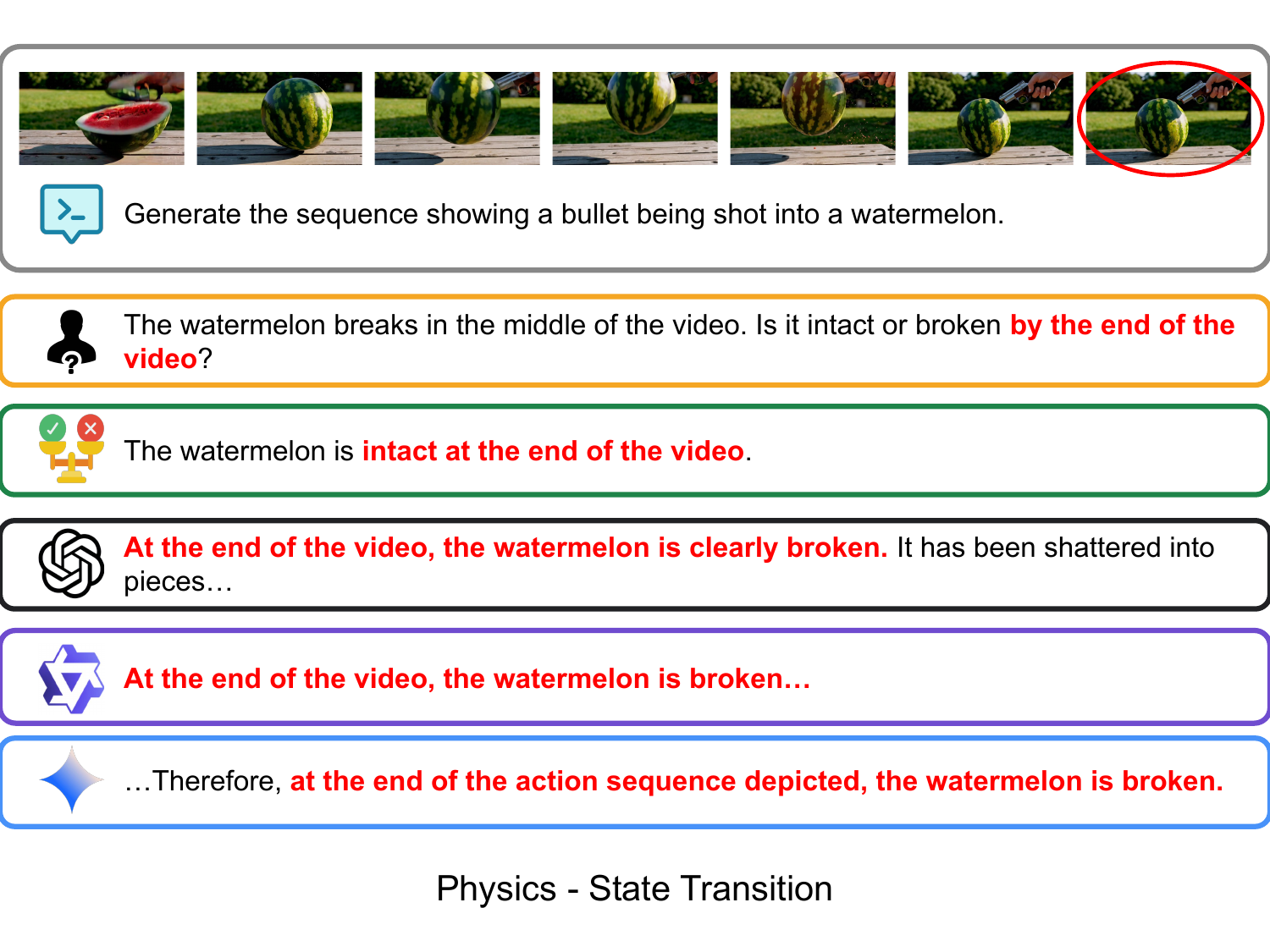}
\caption{\textbf{Hallucination Case from Physics - State Transition (P-ST).}
We show hallucination examples from SOTA MLLM evaluations under the P-ST category. Each case includes the video generation prompt (\textcolor{Gray}{\textbf{Gray}}), key frames from synthetic videos (\textcolor{Gray}{\textbf{Gray}}), questions (\textcolor{Orange}{\textbf{Orange}}), ground truth (\textcolor{Green}{\textbf{Green}}), and model answers from GPT-4o (\textcolor{Black}{\textbf{Black}}), Qwen2.5-VL (\textcolor{Purple}{\textbf{Purple}}), and Gemini-2.5-Pro (\textcolor{Blue}{\textbf{Blue}}), with hallucinations and critical context highlighted in \textcolor{Red}{\textbf{Red}}.}
\label{fig:example-P-ST}
\end{figure}

$$$$

\newpage
\section{Theoretical Problem Formulation}
The motivation of our work comes from the assumption that the language priors within the LLM backbone of the VLMs may interfere with their understanding of synthetic videos. Our goal is to craft a dataset of synthetic videos featuring perceptually obvious violations of common sense and physical laws that require true visual recognition to detect. Let $f_\mathtt{VLM}$, $f_\mathtt{LLM}$ denote the VLM and its LLM backbone, respectively, and $f_\mathtt{Human}$ denote the human expert providing ground truth understanding. $f_\mathtt{VLM}(\textit{video}, \textit{query})$ can take a video-query pair as input, $f_\mathtt{LLM}(\textit{context}, \textit{query})$ can take a text-only context-query pair as input, and $f_\mathtt{Human}(\textit{context}, \textit{query})$ can take multi-modal inputs paired with queries. 
We denote $\mathcal V$ as the set of all contexts within the synthetic video $V$. The context $\mathcal C$ denotes the context being probed during the video understanding process, where $\mathcal Q$ denotes the query probing this context $\mathcal C$. 
We define a mapping function $\mathit{T}(\cdot)$ that transforms a set of contextual elements into a natural language-formulated text for both the query $\mathcal{Q}$ and context $\mathcal{C}$. This mapping can be performed by either humans or LLMs.
We introduce the \textit{contextual distance} \( d[\cdot,\cdot] \) to quantify the semantic divergence between two contexts or texts~\citep{wu2024autohallusion}. When two contexts convey similar or mutually consistent information, \( d \) is small; otherwise, it is large. This metric captures the degree of contextual alignment and can be estimated using \textit{LLM-as-a-Judge} approaches~\citep{zheng2023judgingllmasajudgemtbenchchatbot, kim2024prometheusinducingfinegrainedevaluation, li2024pedantscheapeffectiveinterpretable} or other model-based evaluators.  
In the post-training \textit{human preference alignment} setting, we regard \( f_\mathtt{Human}(\cdot,\cdot) \) as the ground truth and expect both \( f_\mathtt{VLM} \) and \( f_\mathtt{LLM} \) to align with human perception and understanding of the real world. The objective is formulated as:  
\begin{align}
    \max\limits_{V,\mathcal Q,\mathcal C} &~~~~d[f_\mathtt{VLM}(V, \mathit{T}({\mathcal Q})), f_\mathtt{Human}(V, \mathit{T}({\mathcal Q}))] \label{equ:f2s} \\[-1mm]
    \text{s.t.} & ~~~~d[f_\mathtt{LLM}(\mathit{T}({\mathcal C}), \mathit{T}({\mathcal Q})), f_\mathtt{Human}(\mathit{T}({\mathcal C}), \mathit{T}({\mathcal Q}))]\leq \epsilon, \notag \\
    & ~~~~ d[f_\mathtt{Human}(\mathit{T}({\mathcal C}), \mathit{T}({\mathcal Q})), f_\mathtt{Human}(V, \mathit{T}({\mathcal Q}))]\geq \delta, ~ \mathcal C\subseteq \mathcal{V}, \label{equ:f2constrain}
\end{align}
where Equation~(\ref{equ:f2s}) maximizes the contextual distance between the VLM’s output and the human-annotated ground truth for a given synthetic video \( V \) and query \( \mathcal{Q} \). The constraints in~(\ref{equ:f2constrain}) ensure that the language-only model \( f_\mathtt{LLM} \), given the same query \( \mathcal{Q} \) and context \( \mathcal{C} \), remains aligned with human judgment within a tolerance \( \epsilon \), while the video \( V \) introduces human-detectable inconsistencies relative to \( \mathcal{C} \), yielding a contextual distance exceeding a threshold \( \delta \). The context \( \mathcal{C} \) is embedded within \( V \) to preserve coherence.

\newpage
\section{Video Understanding and Evaluation Categorization/Motivation}
\label{app:categorization}
We provide details on specific categorizations of errors video generation models can make.
We draw inspiration from basic video quality evaluation definitions from MVBench~\cite{li2024mvbenchcomprehensivemultimodalvideo} and WorldModelBench~\cite{li2025worldmodelbench} to first organize the current challenges of video generations and evaluations in four basic categories (Figure~\ref{tab:video-eval-compact}).
%
Given the probing target of each question-answering pair and the demand for reasoning abilities or prior knowledge of the LLM backbone to solve the question provided, we divide the question-answering pairs for testing MLLM-as-evaluators into four major categories with sub-categories.

The categorization is to go beyond superficial metrics like frame consistency or resolution by enabling rigorous evaluation through the identification of visual abnormalities across predefined categories. 
To achieve this, we design targeted adversarial questions that expose these anomalies. 
This allows us to assess whether current SOTA MLLMs can effectively detect and interpret such issues, which is an essential step toward scalable and interpretable video evaluation. We further extend these principles to define our video understanding criteria benchmark.

\textbf{Alignment} checks whether the model accurately identifies basic entity details and ensures the video content fully aligns with the prompt without omissions or discrepancies.
%
\begin{itemize} 
    \item \textbf{Entity Counting (A-EC):} Quantifies how many entities are present in the scene.
    \item \textbf{Entity Properties (A-EP):} Focuses on visual features such as color, shape, and texture that define an entity’s appearance. 
    \item \textbf{Entity Recognition and Classification (A-ERAC):} Identifies and categorizes entities based on attributes like shape, color, and texture. 
    \item \textbf{Spatial Relationships (A-SR):} Examines the relative positions of mostly static entities as described in the prompt. 
\end{itemize}

\textbf{Spatial-Temporal Consistency} evaluates whether the model can detect smooth, consistent changes in objects, actions, and viewpoints over time, without abrupt or abnormal transitions in space or time. 
%
\begin{itemize} 
    \item \textbf{Camera Dynamics (SC-CD):} Covers variations in camera movement, angle, and viewpoint.
    \item \textbf{Spatial Dynamics (SC-SD):} Focuses on entity motion, changing positions, and interactions, identifying any inconsistencies or abrupt spatial changes. 
    \item \textbf{Temporal Dynamics (SC-TD):} Tracks changes in entities or scenes over time, including appearance shifts, transformations, and abnormal appearances or disappearances. 
\end{itemize}

\textbf{Common Sense Reasoning} assesses the model’s ability to apply general knowledge and reasoning to detect conflicts between common sense and the visual context, ensuring it interprets the prompt correctly without hallucinating entities or actions. 
\begin{itemize} 
    \item \textbf{Knowledge (CS-K):} Assesses the model’s ability to apply general knowledge of everyday phenomena, including object geometry, layout, and state transitions. 
    \item \textbf{Reasoning (CS-R):} Tests the model’s ability to interpret problem cues—including emotional or environmental hints, and solve them through reflection and chain-of-thought.
\end{itemize}

\textbf{Physics} assesses the model’s ability to detect physical inconsistencies, such as violations of gravity, motion dynamics, or conservation laws, requiring careful reasoning about object properties and movements even if not explicitly stated. 
\begin{itemize} 
    \item \textbf{Conservation (P-C):} Assesses understanding of mass and energy conservation, ensuring entity quantities remain constant unless acted upon by external forces. 
    \item \textbf{Constraints and Properties (P-CAP):} Checks understanding of physical constraints and properties, such as rigid bodies blocking motion or light behavior like reflection. 
    \item \textbf{Motion (P-M):} Evaluates the model's grasp of motion-related physics (like gravity, linear/circular motion, relative movement, and fluid dynamics), spotting inconsistencies or abrupt changes. 
    \item \textbf{State Transition (P-ST):} Tests knowledge of physics-driven state changes, including heat effects, phase transitions, and dynamic interactions. 
\end{itemize}

\newpage
\section{Prompt Templates}
\label{app:prompt_templates}
We provide the prompt templates we use for CoT prompt (Table~\ref{tab:cot_prompt_template}) then generate the final answer (Video-R1-CoT and VideoChat-R1-thinking) and prompt templates for generating answers directly (Table~\ref{tab:direct_answer_prompt_template}).

\begin{table*}[h]
\small
    \setlength{\tabcolsep}{4pt}
    \centering
    \resizebox{0.7\columnwidth}{!}{
    \begin{tabular}{c p{12cm}}
     \toprule
    & \multicolumn{1}{c}{\bf CoT Prompt Template} \\
    \midrule
     \noalign{\vskip 1mm}
     & \texttt{System Prompt: A conversation between User and Assistant. The user asks a question, and the Assistant solves it. The assistant first thinks about the reasoning process in the mind and then provides the user with the answer. The reasoning process and answer are enclosed within <think> </think> and <answer> </answer> tags, respectively, i.e., <think> reasoning process here </think><answer> answer here </answer>
}\\
     \noalign{\vskip 2mm}
     & \texttt{Input: Please think about this question as if you were a human pondering deeply. Engage in an internal dialogue using expressions such as `let me think', `wait', `Hmm', `oh, I see', `let's break it down', etc, or other natural language thought expressions. It is encouraged to include self-reflection or verification in the reasoning process. Provide your detailed reasoning between the <think> </think> tags, and then give your final answer between the <answer> </answer> tags.} \\
     \noalign{\vskip 2mm}
     & \texttt{Question: \{Question\}} \\
     \noalign{\vskip 1mm}
    \bottomrule
    \end{tabular}
    }
    \caption{The prompt template for Video-R1-CoT and VideoChat-R1-thinking to generate answers. This prompt encourages them to first think critically about the video and the question then generate a final answer.} 
    \label{tab:cot_prompt_template}
\end{table*}

\begin{table*}[h]
\small
    \setlength{\tabcolsep}{4pt}
    \centering
    \resizebox{0.7\columnwidth}{!}{
    \begin{tabular}{c p{12cm}}
     \toprule
    & \multicolumn{1}{c}{\bf Direct Answer Prompt Template} \\
    \midrule
     \noalign{\vskip 1mm}
     & \texttt{System Prompt: A conversation between User and Assistant. The user asks a question, and the Assistant solves it. The assistant provide answers within the <answer> </answer> tags: <answer> answer here </answer>
}\\
     \noalign{\vskip 2mm}
     & \texttt{Input: You will be given a video and a question. Please provide an answer to the question based on the video enclosed by <answer> your answer </answer> tags.} \\
     \noalign{\vskip 2mm}
     & \texttt{Question: \{Question\}} \\
     \noalign{\vskip 3mm}
     & \texttt{Answer:} \\
     \noalign{\vskip 1mm}
    \bottomrule
    \end{tabular}
    }
    \caption{Direct answer directly prompts a model to generate the answer without generating additional chain-of-thoughts.} 
    \label{tab:direct_answer_prompt_template}
\end{table*}

\begin{table*}[h]
\small
    \setlength{\tabcolsep}{4pt}
    \centering
    \resizebox{0.7\columnwidth}{!}{
    \begin{tabular}{c p{12cm}}
     \toprule
    & \multicolumn{1}{c}{\bf LLM-as-A-Judge Prompt Template} \\
    \midrule
     \noalign{\vskip 1mm}
     & \texttt{You will be given a question, a reference answer, and a predicted response. You task is to judge the correctness of the predicted response. If the predicted response is correct, please answer "correct". If the predicted response is incorrect, please answer "incorrect". Please strictly follow the output format below.} \\
     \noalign{\vskip 2mm}
     & \texttt{OUTPUT FORMAT:} \\
     \noalign{\vskip 2mm}
     & \texttt{Judgment: YOUR JUDGMENT} \\
     \noalign{\vskip 2mm}
     & \texttt{Question: \{Question\}} \\
     \noalign{\vskip 3mm}
     & \texttt{Reference Answer: \{Reference Answer\}} \\
     \noalign{\vskip 2mm}
     & \texttt{Predicted Answer: \{Predicted Response\}} \\
     \noalign{\vskip 2mm}
     & \texttt{YOUR OUTPUT:} \\
     \noalign{\vskip 1mm}
    \bottomrule
    \end{tabular}
    }
    \caption{LLM-as-a-judge prompt template.} 
    \label{tab:llm_judge_prompt_template}
\end{table*}

\newpage
\section{Categorization Breakdown Results}
We provide a qualitative breakdown of results in multiple radar charts across fine-grained categories for the evaluated baselines, serving as supplementary analysis to Table~\ref{tab:acc_by_category}.
\begin{figure}[htbp]
\centering
\begin{minipage}{0.24\linewidth}
    \centering
    \includegraphics[width=\linewidth]{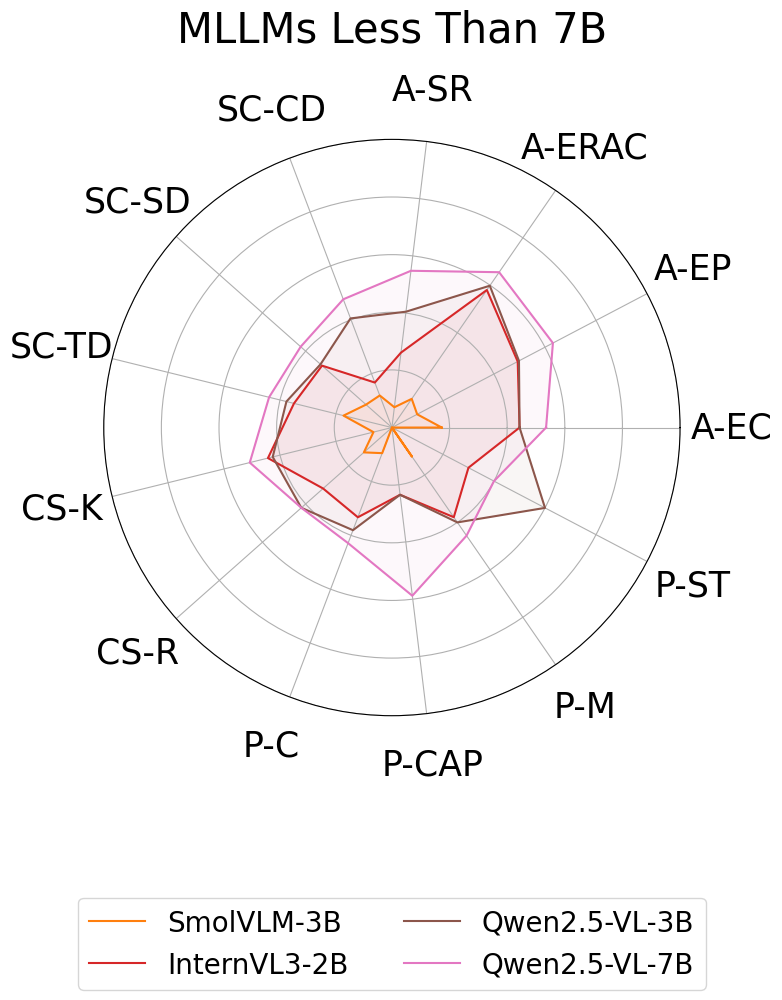}
    \label{fig:radar-1-1}
\end{minipage}
\hfill
\begin{minipage}{0.24\linewidth}
    \centering
    \includegraphics[width=\linewidth]{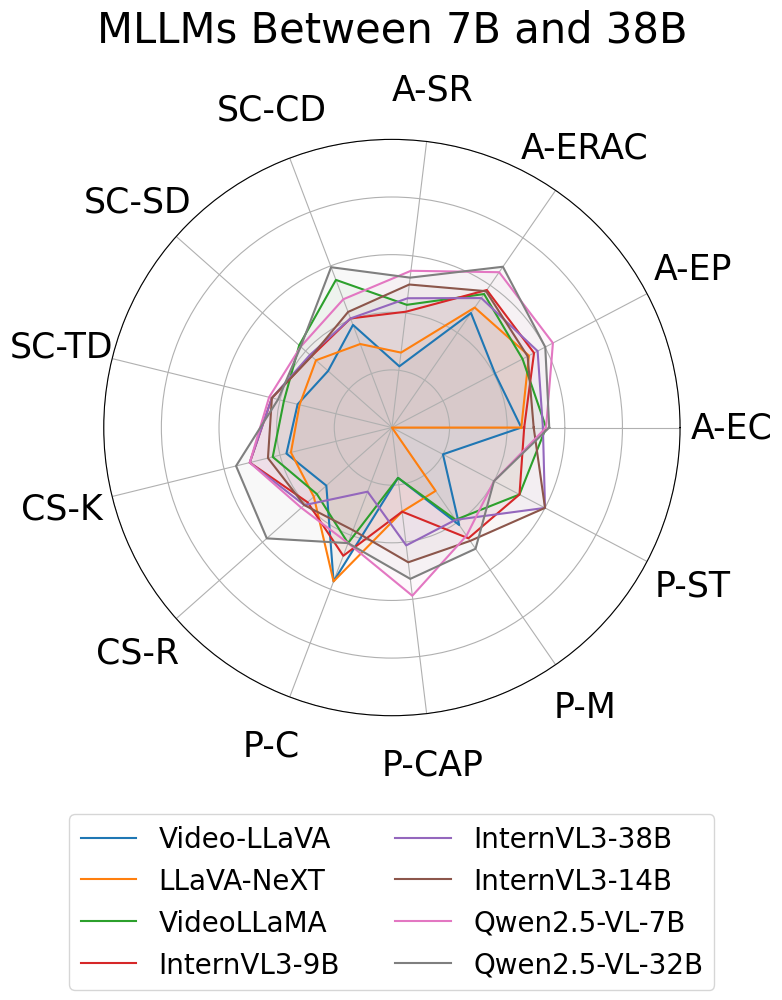}
    \label{fig:radar-1-2}
\end{minipage}
\hfill
\begin{minipage}{0.24\linewidth}
    \centering
    \includegraphics[width=\linewidth]{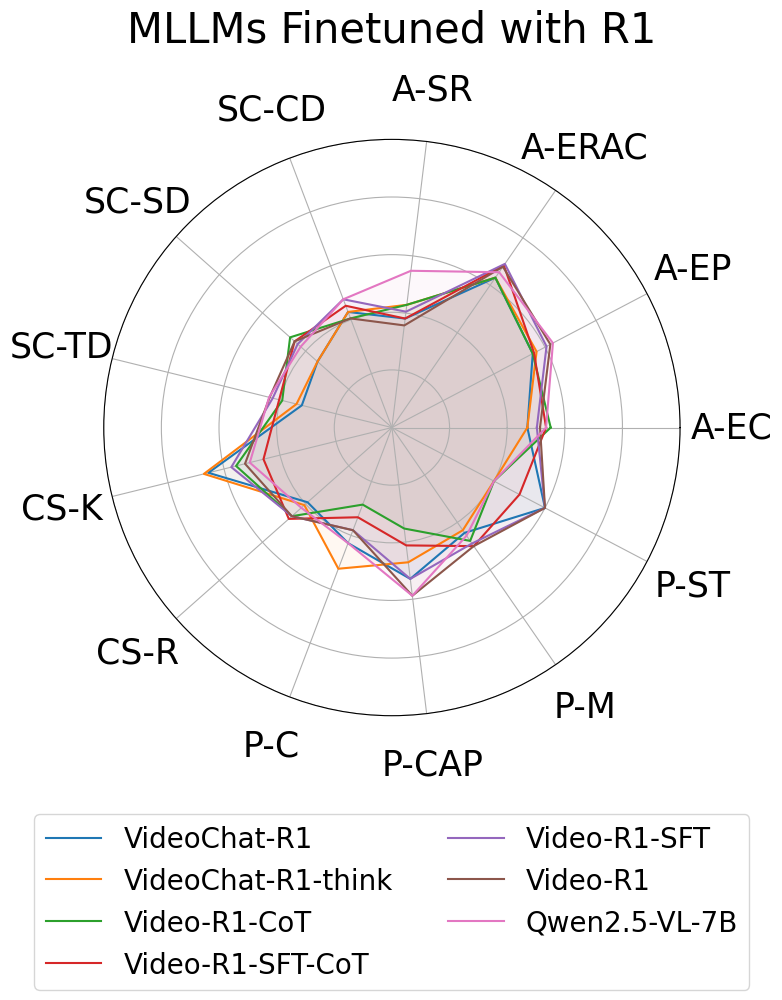}
    \label{fig:radar-1-3}
\end{minipage}
\hfill
\begin{minipage}{0.24\linewidth}
    \centering
    \includegraphics[width=\linewidth]{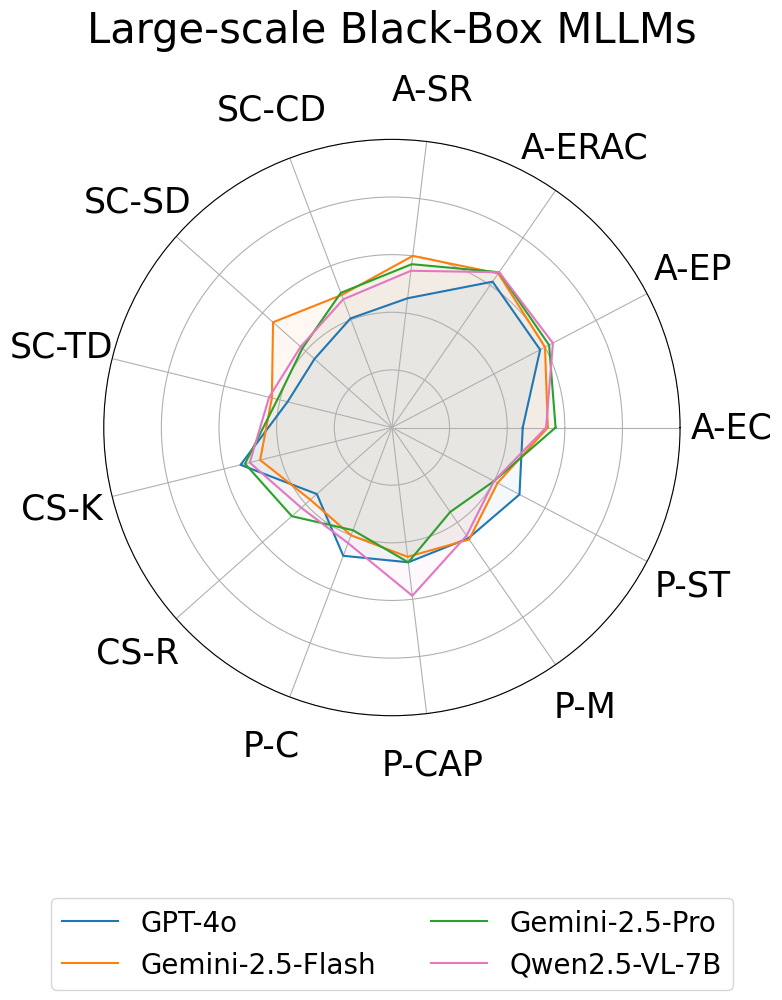}
    \label{fig:radar-1-4}
\end{minipage}
\caption{\textbf{SOTA VLM Evaluation on \ours{} Across Sub-Categories.} We evaluate SOTA VLMs on \ours{}, with results broken down by sub-category. From left to right, we show: (a) models under 7B parameters; (b) models between 7B–38B; (c) R1 fine-tuned models; and (d) large black-box VLMs. While many perform well on alignment tasks, they remain prone to hallucinations in reasoning-heavy tasks, with notably weaker performance on physics and commonsense reasoning.}
\label{fig:acc_by_category_radar}
\vspace{-15pt}
\end{figure}

\begin{figure}[htbp]
\centering
\begin{minipage}{0.25\linewidth}
    \centering
    \includegraphics[width=\linewidth]{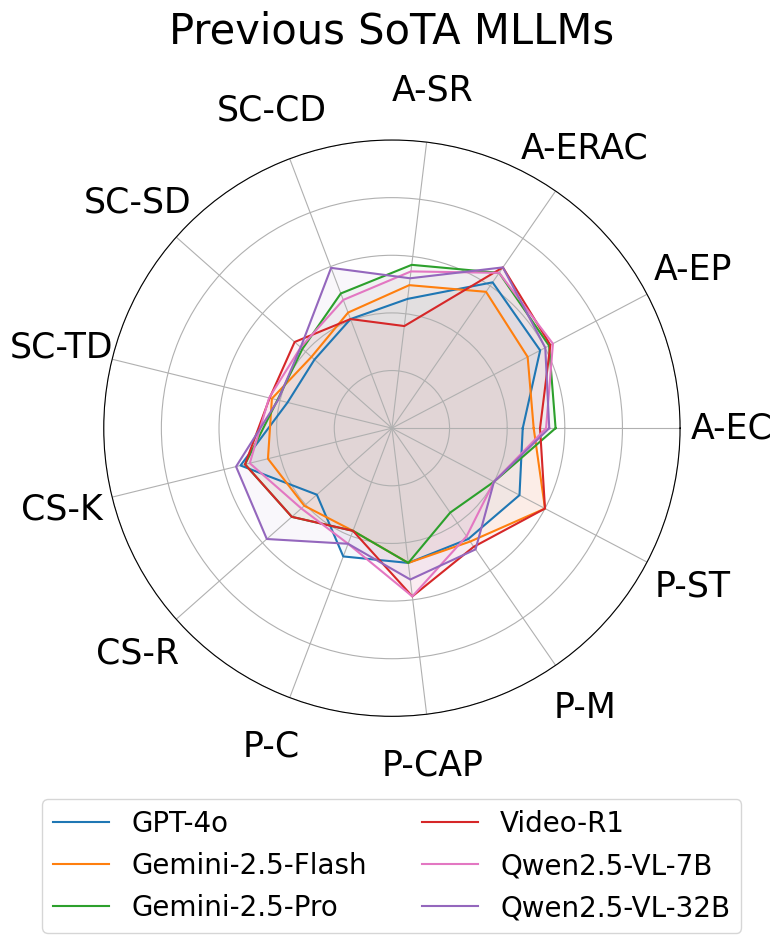}
    \label{fig:radar-1.1-1}
\end{minipage}
\hfill
\begin{minipage}{0.25\linewidth}
    \centering
    \includegraphics[width=\linewidth]{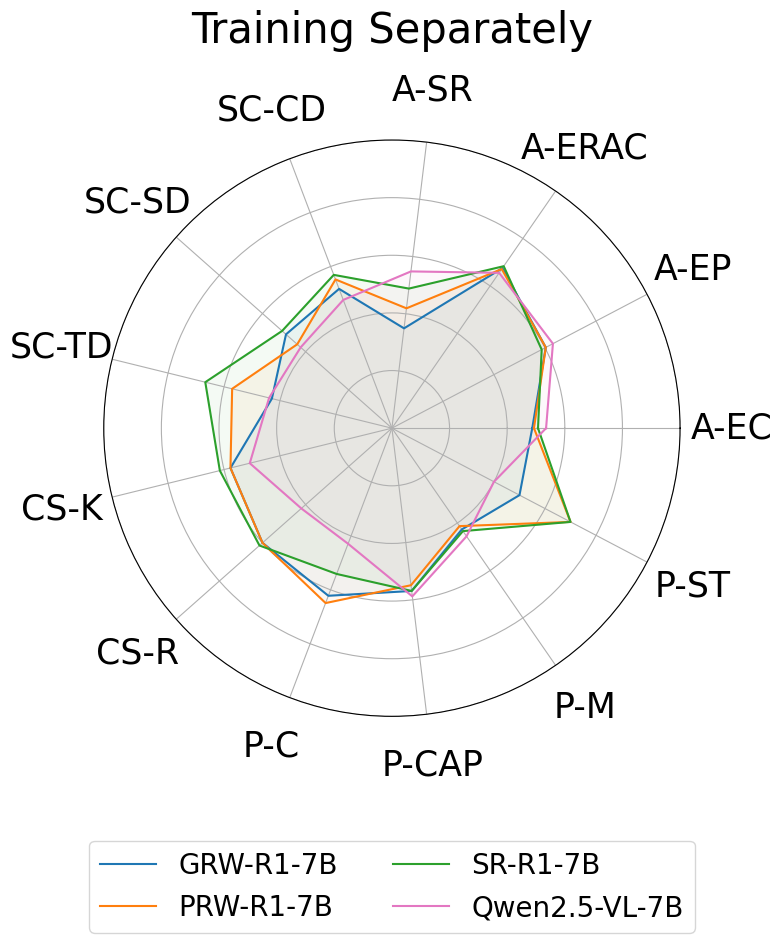}
    \label{fig:radar-1.1-2}
\end{minipage}
\hfill
\begin{minipage}{0.25\linewidth}
    \centering
    \includegraphics[width=\linewidth]{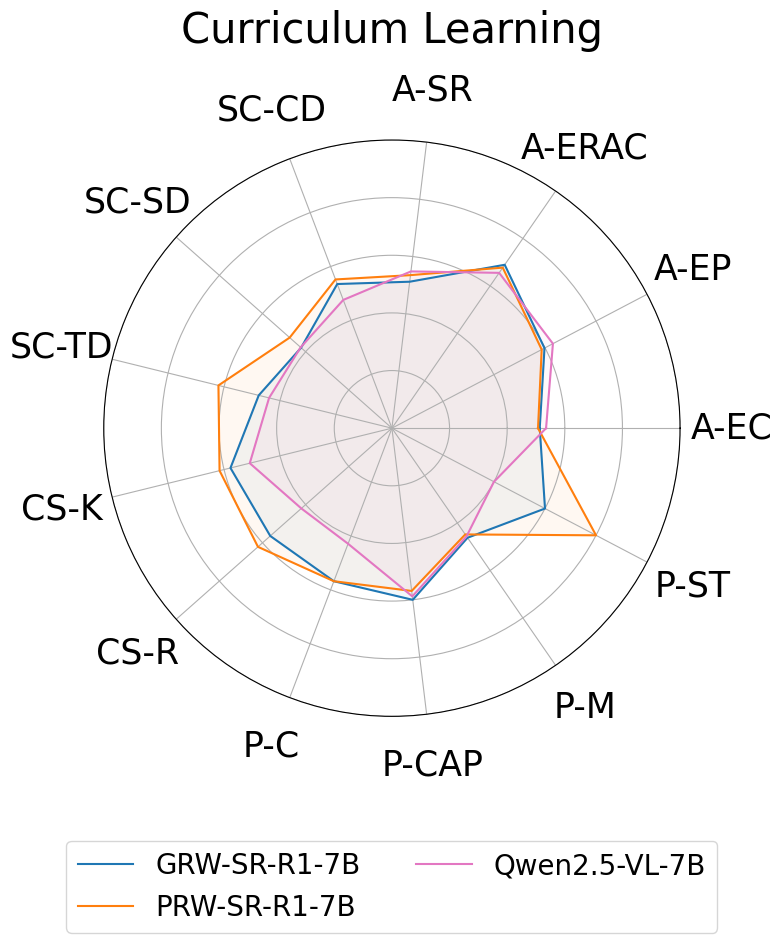}
    \label{fig:radar-1.1-3}
\end{minipage}
\caption{\textbf{Evaluation Breakdown of Fine-Tuned Models.} We show results for (a) previous SOTA VLMs, (b) models fine-tuned on sub-datasets, and (c) models fine-tuned on the full dataset via curriculum learning. Compared to the baseline (Qwen2.5-VL-7B), RFT on commonsense and physics data improves models' reasoning and overall performance in synthetic video understanding.}
\label{fig:radar_rft}
\vspace{-15pt}
\end{figure}

\newpage
\section{Common Sense and Video-dependent Question-Answering}
Our benchmark, \ours{}, is designed to evaluate MLLMs' abilities to detect abnormalities in synthetic videos—a task often confounded by hallucinations stemming from commonsense or physical knowledge embedded in their language priors. This section breaks down model performance across question types in \ours{}, including:
\begin{itemize}
    \item \textbf{Common Sense-only Questions:} These can be answered using language priors alone, without relying on video input. \textit{e.g., What typically happens when a bullet hits a watermelon?} (Answer: \textit{It explodes into pieces.})
    \item \textbf{Counterintuitive Questions:} Target counterfactual contexts in synthetic videos, testing whether MLLMs can recognize visually implausible phenomena. \textit{e.g. In the video (Sora), the watermelon breaks in the middle of the video. Is it intact or broken at the end?} (Answer:  \textit{It's intact.}) (Figure~\ref{fig:teaser})
    \item \textbf{Critical Thinking Questions:} Open-ended questions that ask whether MLLMs can identify abnormalities in synthetic videos, evaluating their visual reasoning. \textit{e.g. What is unusual in this video (Sora)?} (Answer: \textit{The watermelon explodes, then reassembles.}) (Figure~\ref{fig:teaser})
\end{itemize}

while the latter two types of questions must be answered with video inputs, so that we denote them as video-dependent questions.

\begin{table}[ht]
\centering
\tiny
\setlength{\tabcolsep}{12pt} 
\begin{tabular}{l|c|cc|c}
\toprule
\multirow{3}{*}{Model} & \multirow{3}{*}{Common Sense-only} & \multicolumn{2}{c|}{Video-dependent} &  \multirow{3}{*}{Overall}\\ 
\cmidrule(lr){3-4}
 &    & Counterintuitive  & Critical Thinking &  \\
\midrule
\texttt{GPT-4o} & 100.0 & 46.8 & 15.0 & 45.5 \\
\texttt{InternVL3-14B} & 100.0 & 48.2 & 10.0 & 46.7\\
\texttt{Gemini-2.5-Pro} & 100.0 & 50.2 & 23.3 & 49.8\\
\texttt{Video-R1} & 100.0 & 52.3 & 16.7 & 50.8 \\
\texttt{Qwen2.5-VL-7B} & 100.0 & 53.1 & 10.0 & 51.0\\
\texttt{Qwen2.5-VL-32B} & 100.0 & 52.5 & 13.3 & 51.4\\
\bottomrule
\end{tabular}
\vspace{6 pt}
\caption{\textbf{Common Sense and Video-dependent QA over \ours{}.} We divide \ours{} into multiple categories over the question types: \textbf{(a) Common Sense-only Questions,} answerable via language priors without video inputs; \textbf{(b) Counterintuitive Questions,} probing MLLMs' abilities in detecting counterintuitive phenomena; and \textbf{(c) Critical Thinking Questions,} assessing MLLMs' ability to detect abnormalities in synthetic videos.
}
\label{tab:commonsense}
\vspace{-10pt}
\end{table}

In Table~\ref{tab:commonsense}, we show the evaluation breakdown by question type for six SOTA MLLMs. All models reach 100.0\% accuracy on commonsense-only questions, indicating strong grounding in pre-trained knowledge. However, performance drops on counterintuitive questions (all below 55\%) and further on critical thinking questions, where no model exceeds 25\% accuracy, revealing major limitations in detecting and reasoning about abnormalities based on physics and commonsense.

Gemini-2.5-Pro performs best on critical thinking (23.3\%), followed by Video-R1 (16.7\%), suggesting some benefit from CoT prompting. However, CoT remains unreliable under language prior bias and does not consistently improve abnormality detection. Enhancing MLLMs' critical thinking for such tasks remains an open challenge.

Counterintuitive questions typically include contextual hints, helping models locate anomalies. In contrast, critical thinking questions are open-ended, requiring models to identify and reason about abnormalities unaided, making them more vulnerable to hallucinations when their video understanding is incomplete.

\end{document}